\documentclass{article} 
\usepackage[preprint]{colm2025_conference}

\usepackage{float}
\usepackage{multirow}
\usepackage{amsmath}
\usepackage{tcolorbox}
\usepackage{listings}
\usepackage{lmodern}  
\usepackage{microtype}
\usepackage{hyperref}
\usepackage{url}
\usepackage{booktabs}
\usepackage{graphicx}
\usepackage{lineno}
\usepackage{colortbl}
\usepackage{subcaption}
\definecolor{lightblue}{RGB}{173,216,230}
\definecolor{lightyellow}{RGB}{255,255,182}
\usepackage[inline]{enumitem}
\usepackage{cleveref}

\definecolor{darkblue}{rgb}{0, 0, 0.5}
\hypersetup{colorlinks=true, citecolor=darkblue, linkcolor=darkblue, urlcolor=darkblue}

\title{A Llama walks into the 'Bar': Efficient Supervised Fine-Tuning for Legal Reasoning in the Multi-state Bar Exam}

\author{Rean Fernandes,$^1$ André Biedenkapp,$^1$ Frank Hutter,$^{2,1}$ Noor Awad$^1$\\
	$^1$ Albert Ludwigs University Freiburg, Germany\\
	$^2$ ELLIS Institute Tübingen, Germany\\
	\texttt{reanclive@gmail.com, \{biedenka,fh,awad\}@cs.uni-freiburg.de}
}

\begin{document}

\maketitle

\begin{abstract}
Legal reasoning tasks present unique challenges for large language models (LLMs) due to the complexity of domain-specific knowledge and reasoning processes. This paper investigates how effectively smaller language models (Llama 2 7B and Llama 3 8B) can be fine-tuned with a limited dataset of 1,514 Multi-state Bar Examination (MBE) questions to improve legal question answering accuracy. We evaluate these models on the 2022 MBE questions licensed from JD Advising, the same dataset used in the 'GPT-4 passes the Bar exam' study. Our methodology involves collecting approximately 200 questions per legal domain across 7 domains. We distill the dataset using Llama 3 (70B) to transform explanations into a structured IRAC (Issue, Rule, Application, Conclusion) format as a guided reasoning process to see if it results in better performance over the non-distilled dataset. We compare the non-fine-tuned models against their supervised fine-tuned (SFT) counterparts, trained for different sample sizes per domain, to study the effect on accuracy and prompt adherence. We also analyse option selection biases and their mitigation following SFT. In addition, we consolidate the performance across multiple variables: prompt type (few-shot vs zero-shot), answer ordering (chosen-option first vs generated-explanation first), response format (Numbered list vs Markdown vs JSON), and different decoding temperatures. Our findings show that domain-specific SFT helps some model configurations achieve close to human baseline performance, despite limited computational resources and a relatively small dataset. We release both the gathered SFT dataset and the family of Supervised Fine-tuned (SFT) adapters optimised for MBE performance. This establishes a practical lower bound on resources needed towards achieving effective legal question answering in smaller LLMs.
\end{abstract}

\section{Introduction}
Large language models (LLMs) have demonstrated remarkable abilities in complex reasoning tasks across various domains. OpenAI's GPT-4 achieved a significant milestone in the legal field when it passed the Uniform Bar Examination (UBE) in a zero-shot setting \citep{Katz+2023}, reportedly outperforming the passing threshold in all jurisdictions and surpassing human test-takers. This is a considerable achievement as the the bar exam  represents the final hurdle that aspiring lawyers must clear to be licensed to practice law. This exam requires factual knowledge and analytical reasoning making it a good benchmark for evaluating legal reasoning capabilities in LLMs.

Larger commercial models like GPT-4 and Claude are essentially black boxes, with limited transparency regarding their model size and pretraining data was used. The capabilities of such models in a zero-shot setting might not be so surprising, given the huge capital and the compute used for their training. This is why, we try to investigate the following question in our paper: \textbf{Can we achieve similar performance with smaller, open-weight models, with a limited dataset of questions, capable of being run on consumer grade hardware?}

Our work tries to answer this question by applying supervised fine-tuning (SFT) to the Llama family of models, specifically Llama 2 7B and Llama 3 8B, fine-tuned to answer the Multistate Bar Exam (MBE). We perform SFT using a limited dataset of $1514$ questions from previous year MBE exams, spread across the $7$ legal domains tested in the exam. Our evaluation benchmark uses the same dataset that was used by \citet{Katz+2023}, allowing us to remain faithful to the comparison method.

We aim to determine whether there is a lower bound on task-specific samples required to achieve comparable performance with bigger SOTA models like GPT-4. Through detailed empirical analysis, we study performance impacts across multiple generation-based parameters. This approach helps consolidate our findings on model performance after fine-tuning.

In addition to our work, our contributions to the open source community are as follows:
\begin{enumerate}
    \item We release the curated fine-tuning dataset of 1,514 MBE questions meticulously gathered from online study materials, releasing both the un-distilled dataset and the dataset processed into structured reasoning formats. \footnote{The datasets are available at \url{https://huggingface.co/datasets/HolySaint/MBE-exam-questions}}
    \item We present the family of fine-tuned model adapters for use with Llama 2 7B and Llama 3 8B to answer the MBE. \footnote{All the trained SFT adapters are available at \url{https://huggingface.co/HolySaint/bar-Llama-adapters}}
\end{enumerate}

We restrict our work to Supervised Fine-Tuning (SFT) with quantized LoRA \citep{dettmers2023qloraefficientfinetuningquantized}, deliberately excluding reinforcement learning-based approaches such as PPO \citep{schulman2017proximalpolicyoptimizationalgorithms} and GRPO \citep{shao2024deepseekmathpushinglimitsmathematical} which utilize proximal policy optimization techniques, and Direct Preference Optimization (DPO) \citep{NEURIPS2023_a85b405e}. This decision allows us to quantify performance improvements from a more fundamental baseline perspective.
\section{Related works}
\subsection{GPT-4 and its performance on the Unified Bar Exam}
Recent work by \citet{Katz+2023} reported that GPT-4 achieved strong performance on the Uniform Bar Exam (UBE), claiming near-90th percentile results. However, subsequent analysis by \citet{martinez2023gpt4barperformance} identified several methodological issues that affect these performance metrics. Martinez's reanalysis found that GPT-4 achieved only the 68th percentile among July test-takers and only the 45th percentile among qualified attorneys when evaluated against more representative testing populations. GPT-4 performed notably better on multiple-choice sections (MBE) than on written portions (MEE and MPT), where its performance dropped to the 15th percentile among qualified attorneys. We talk more about these findings in appendix section \ref{app:gpt4-bar-exam}.

Given these methodological questions surrounding GPT-4's bar exam performance, particularly regarding test evaluation methodologies and the representativeness of reported metrics, our work takes a more systematic approach. While we also cannot make any guarantees about the pretraining data used for the models in our study, we seek to establish the baseline by first running inference on the models without any SFT, to see whether they have already seen the test data, and then compare their performance after SFT.

\subsection{SFT Work for legal domains}
\textbf{SaulLM} \citep{colombo2024saullm54bsaullm141bscaling,colombo2024saullm7bpioneeringlargelanguage} is a family of LLMs built upon the open source Mistral \citep{jiang2023mistral7b} and Mixtral \citep{jiang2024mixtralexperts} models, specifically adapted for legal applications. The SaulLM developers curated an intensive corpus of 30 billion tokens from various legal sources and conducted both continued pretraining and instruction fine-tuning. They reported state-of-the-art performance on LegalBench, a collaborative benchmark for testing legal reasoning in language models \citep{guha2023legalbenchcollaborativelybuiltbenchmark}.
While LegalBench encompasses multiple legal tasks, we found these tasks too specific for our goal of evaluating performance on the MBE. Therefore, we opted to curate our own custom SFT dataset specifically geared toward answering MBE questions. Future work will include evaluating our fine-tuned model adapters on the LegalBench for a more comprehensive legal application study. The original SaulLM continuous pretraining was performed on 384 AMD MI250 (128GB) GPUs, with SFT and preference optimization conducted on 64 of the same hardware units.

\textbf{DISC-LawLLM} \citep{yue2023disclawllmfinetuninglargelanguage} is a comprehensive system of LLMs that have been continuously pretrained and then fine-tuned with a curated dataset consisting of questions from Chinese legal exams. They also use both retrieval augmented generations to mitigate hallucinations, and rely on a more capable model (GPT-4) to subjectively evaluate the response of these fine-tuned LLMs on the same exams, and report better performance over GPT-3.5 evaluated in a zero shot setting. They used 8 x NVIDIA A800 (40GB) GPUs for the continuous pretraining and subsequent fine-tuning. 

The key takeaway from these studies is that the authors invested considerable effort in collecting and distilling extensive datasets into high-quality corpora. They engineered complex pipelines to both perform thorough evaluations and reduce hallucinations, while utilizing computational resources on the scale of multiple high-end GPUs. The substantial effort required and the computational costs make reproducing such work challenging. This challenge motivates our research to establish a lower bound on what could be achievable using a single GPU and a relatively small SFT dataset.

\section{Approach}
In this section, we detail our dataset collection process, highlighting how we used a more capable model to distill and extract explanations from the original data. We then analyze the key parameters influencing model prompting and subsequent response generation. Subsequently, we describe our fine-tuning and inference pipeline, along with the computational resources employed. To supplement this section we add more details in appendix section \ref{section:approach_appendix}.

\subsection{Distilling the dataset into the IRAC format}
Our training dataset consists of previous bar exam questions gathered from online study materials. These practice questions differ from actual exams since new and unique questions are created each year. We extensively verified that there was no overlap between our curated SFT dataset and the test set licensed from JD Advising.\footnote{We will not be releasing these test questions as they are officially licensed. Interested readers can source the dataset from \url{https://jdadvising.com/product/200-mbe-question-exam-2022/}.}
Every question in the training set, contains the question body, four possible options, and the explanation justifying the correct option. A possible challenge in our dataset is that the explanations are predominantly written in an informal and unstructured tone. These explanations were possibly written to convey crucial facts to students already familiar with examination patterns to save the time taken to analyse each answer while studying. We hypothesized that this informal nature might impede the model's ability to reason in a structured, test-oriented manner. To investigate this impact, we created an alternative SFT dataset that is identical to the original dataset except for the reformatted explanations.
\paragraph{A framework to answer Legal questions}We adopted the \emph{Issue, Rule, Application, and Conclusion (IRAC)} framework, a standard methodology in answering the bar that provides a systematic approach to analyzing the questions. The IRAC structure helps identify the legal issue at hand, state the relevant legal rule, apply that rule to the facts of the case, and draw a conclusion based on this application \citep{JDAdvising2021}. We consider this framework because it mirrors the analytical process expected of test-takers. To implement this reformatting, we utilized a quantized Llama-3 70B model \citep{dubey2024llama3herdmodels}. We selected this model based on the \citet{kaplan2020scalinglawsneurallanguage} study that larger model size correlates with better performance, making it suitable for generating higher quality structured explanations.
Using this distillation pipeline, we processed our original dataset to create two parallel versions: one containing the original raw explanations and another with the same explanations restructured into the IRAC format. We fine-tuned our models on both datasets to compare the effect of explanation structure on performance. We show an example of the explanation before and after distillation in \autoref{section:distilled_explanation}.

\subsection{Expected response from the LLM}
For any instance of the LLM that we run inference on, whether fine-tuned or not, we expect the following three entities in the response, which will be extracted and considered as the response of the model. 
\begin{enumerate}[label=, leftmargin=20pt, labelindent=\parindent,
listparindent=\parindent, labelwidth=0pt]
    \item \textbf{Chosen domain}: Any one of the 7 domains tested for in the MBE are considered as valid response. 
    \item \textbf{Chosen Option}: Can only be either A or B or C or D. 
    \item \textbf{Explanation}: Supporting or deducing the selected option.
\end{enumerate}
In accordance with how the actual MBE is held, \emph{we consider a response to a question as correct, when the chosen option matches the ground truth label}. We do not consider the chosen domain or the generated explanation to determine correctness, and only take them to be a guidance to the model in its generation.

\subsection{Generation parameters}

We systematically explore how different prompt configurations affect LLM performance on legal reasoning tasks. These parameters influence both how the model structures its responses and its accuracy in answering questions. For our fine-tuned models, these parameters shape the system prompts and format the training datasets. For untrained baseline models, we use the same system prompts to ensure fair comparison across all parameter configurations.

\subsubsection{Response Format and Type}
For \textbf{response format}, we test three structures:
\begin{itemize}[noitemsep,topsep=0pt,parsep=0pt,partopsep=0pt]
    \item \textbf{JSON}: Structured with curly braces that can be directly parsed using JSON libraries, eliminating the need for regex or NLP methods.
    \item \textbf{Markdown}: Using hash symbols and formatting conventions.
    \item \textbf{Numbered list}: Simple sequential organization.
\end{itemize}
This allows us to evaluate whether format complexity affects accuracy and which formats enable the most reliable parsing of model outputs.

For \textbf{response type}, we compare two approaches:
\begin{itemize}[noitemsep,topsep=0pt,parsep=0pt,partopsep=0pt]
    \item \textbf{Fact-first generation}: The model produces analysis before selecting an answer, mirroring sequential reasoning.
    \item \textbf{Answer-first generation}: The model selects an answer before providing justification.
\end{itemize}
These approaches represent different cognitive strategies for solving legal problems.
\subsubsection{Explanation type and Prompt type}
We examine whether structured explanations outperform unstructured ones:
\begin{itemize}[noitemsep,topsep=0pt,parsep=0pt,partopsep=0pt]
    \item \textbf{Structured}: Explanations organized into Issue, Rule, Application, and Conclusion
    \item \textbf{Unstructured}: Free-form explanations without explicit organizational elements
\end{itemize}

Additionally, we assess the impact of providing examples in the prompt:
\begin{itemize}[noitemsep,topsep=0pt,parsep=0pt,partopsep=0pt]
    \item \textbf{Zero-shot}: No examples provided before the test question
    \item \textbf{One-shot}: One example question and solution provided before the test question
\end{itemize}
This allows us to measure how much the models benefit from contextual examples in the prompt.

\subsection{Fine-tuning and inference}
We use Q-LoRa \citep{dettmers2023qloraefficientfinetuningquantized} owing to its lower memory requirement as compared to LoRa \citep{hu2021loralowrankadaptationlarge}, allowing us to fine-tune with larger batch sizes. Our exploratory findings with LoRa did not seem to show better performance, given the same hardware constraint. We format the fine-tuning dataset according to the selected generation parameters, then conduct fine-tuning as a text completion task where the LLM learns to predict the next token.
Each adapter is trained using a single NVIDIA Tesla V100 PCIe 32 GB GPU. For details of our setup we refer to \autoref{section:hyperparams}.
During inference, we process each question from the test dataset independently, without any caching between questions, to ensure complete isolation and prevent any potential context leakage across questions.
One run through the pipeline involves the following steps: 
\begin{enumerate}[itemsep=1pt,parsep=0pt]
    \item Selection of Generation parameters.
    \item Inference on the untrained models and recording of responses.
    \item Selection of SFT dataset (structured or unstructured) and formatting according to selected generation parameters. 
    \item Fine-tuning using QLoRa resulting in trained model adapter.
    \item Loading the adapter with the full model, and inference on the test sets.
    \item Parsing and grading of responses from all the inferences. 
\end{enumerate}
Thus, for any selected parameter configuration, we first establish the baseline with the untrained model, and then evaluate the fine-tuned models and record the accuracy metrics after parsing. 

\section{Experiments and Results}
In this section, we begin by investigating the immediate benefit that fine-tuning provides to the model in terms of its accuracy, and also its adherence to the expected generation format. We then analyse how the inherent bias in terms of preference for certain options is mitigated by the fine-tuning, and then study the impact of fine-tuning on the specific legal domains. 

\subsection{Impact of Fine-Tuning on Model Performance}
\begin{figure}[h]
    \centering
    \begin{subfigure}{0.48\textwidth}
        \centering
        \includegraphics[width=\linewidth]{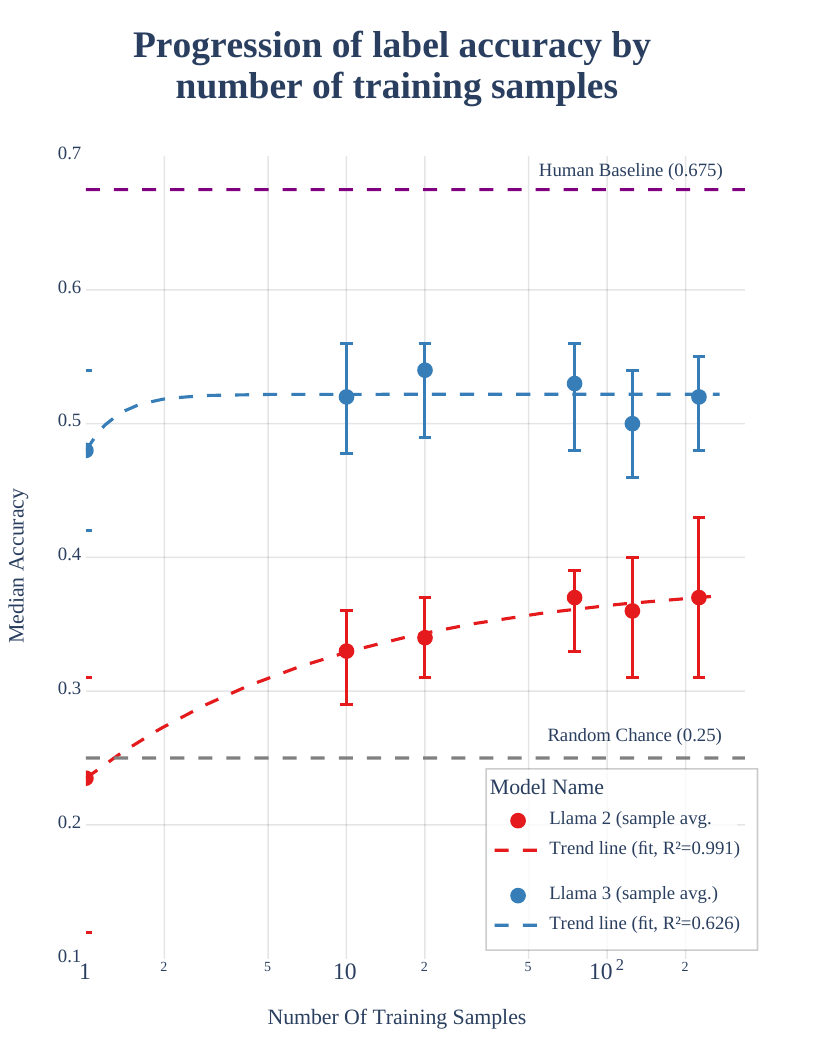}
        \caption{}
        \label{subfig:learning_curve_comparison}
    \end{subfigure}
    \hfill
   \begin{subfigure}{0.48\textwidth}
        \centering
        \includegraphics[width=\linewidth]{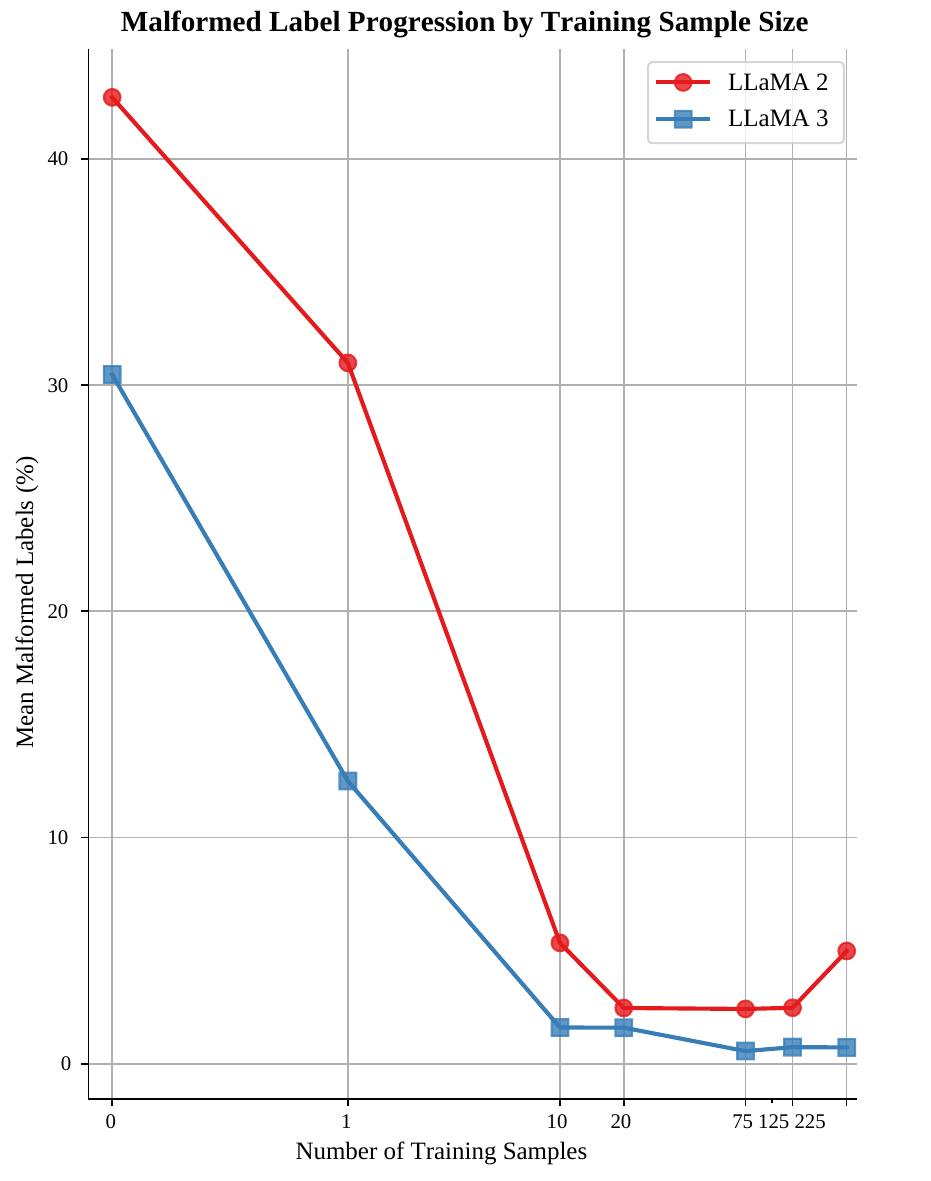}
        \caption{}
        \label{subfig:parsing_failures_comparison}
    \end{subfigure}
    
    \caption[Comparison of parsing failures and learning curves]{\autoref{subfig:learning_curve_comparison} shows learning curves comparing Llama 2 and Llama 3 model performance as a function of training samples. \autoref{subfig:parsing_failures_comparison} shows the reduction in parsing failures with increased fine-tuning samples, demonstrating that models rapidly adapt to the required response format, even with minimal training.}
    \label{fig:combined_comparison}
\end{figure}

\autoref{subfig:learning_curve_comparison} presents the learning curves comparing Llama 2 and Llama 3 models performance as a function of training samples.
The number of samples used in fine-tuning significantly impacts model performance. Llama 3 demonstrates higher initial accuracy ($0.48$) with relatively modest improvement over training samples (R$^2=0.626$), while Llama 2 starts near random chance but exhibits steady, predictable gains (R$^2=0.991$). Both models plateau below the human baseline performance ($0.675$), with Llama 3 achieving maximum accuracy ($0.54$) after only $20$ samples, compared to Llama 2's peak ($0.37$) at $225$ samples.
While both models fail to reach the passing threshold, we attribute this primarily to the limitations of the model architecture sizes used during inference.

Llama 2 demonstrates strong agreement with the power law (R$^2=0.991$), suggesting that increasing the training samples would yield proportionally better performance. In contrast, Llama 3's higher initial capabilities can be attributed to the instruct fine-tuned model's superior ability to follow prompts. Nevertheless, both models show clear performance improvements with additional training samples. \autoref{table:model-accuracy-by-samples} shows the best and second best accuracies, further highlighting that Llama 3 plateaus in performance beyond $20$ samples per domain in the training set.

\begin{table}[h]
\begin{center}
\begin{tabular}{lcccccccc}
\toprule
\multirow{2}{*}{\textbf{Train set sample size}} & \multicolumn{4}{c}{\textbf{Llama 2}} & \multicolumn{4}{c}{\textbf{Llama 3}} \\
\cmidrule(lr){2-5} \cmidrule(lr){6-9}
& \textbf{Mean} & \textbf{Std} & \textbf{Median} & \textbf{Max} & \textbf{Mean} & \textbf{Std} & \textbf{Median} & \textbf{Max} \\
\midrule
Untrained baseline& $0.185$ & $0.144$ & $0.220$ & $0.42$ & $0.358$ & $0.206$ & $0.45$ & $0.62$ \\
$1$ & $0.219$ & $0.111$ & $0.235$ & $0.43$ & $0.461$ & $0.112$ & $0.48$ & $0.63$ \\
$10$ & $0.317$ & $0.069$ & $0.330$ & $0.44$ & $0.516$ & $0.057$ & $0.52$ & $0.65$ \\
$20$ & $0.341$ & $0.049$ & $0.340$ & $0.46$ & \cellcolor{blue!25}$0.525$ & \cellcolor{blue!25}$0.049$ & \cellcolor{blue!25}$0.54$ & \cellcolor{blue!25}$0.67$ \\
$75$ & \cellcolor{lightyellow}$0.359$ & \cellcolor{lightyellow}$0.053$ & \cellcolor{lightyellow}$0.370$ & \cellcolor{lightyellow}$0.47$ & \cellcolor{lightyellow}$0.520$ & \cellcolor{lightyellow}$0.050$ & \cellcolor{lightyellow}$0.53$ & \cellcolor{lightyellow}$0.62$ \\
$125$ & $0.355$ & $0.056$ & $0.360$ & $0.46$ & $0.501$ & $0.054$ & $0.50$ & $0.64$ \\
$225$ & \cellcolor{blue!25}$0.368$ & \cellcolor{blue!25}$0.066$ & \cellcolor{blue!25}$0.370$ & \cellcolor{blue!25}$0.49$ & $0.516$ & $0.048$ & $0.52$ & $0.62$ \\
\bottomrule
\end{tabular}
\end{center}
\caption{Performance comparison of Llama 2 and Llama 3 models across different training sample sizes. Best/2nd-best performing configurations are highlighted in \colorbox{blue!25}{blue}/\colorbox{lightyellow}{yellow}.}\label{table:model-accuracy-by-samples}
\end{table}

\subsection{Impact of Fine-Tuning on Response Parsing}
Reliable extraction of model responses constitutes an essential component of the inference pipeline. \autoref{subfig:parsing_failures_comparison} shows that fine-tuning dramatically improves response format adherence. With untrained models, Llama 2 exhibits a malformed response rate of 42.72\%, while Llama 3 shows a lower rate of 30.47\%. Remarkably, fine-tuning with just a single sample reduces Llama 2's error rate to 30.98\% (a 27\% relative improvement) and Llama 3's rate to 12.50\% (a 59\% relative improvement).This improvement continues with additional training samples, with both models decreasing parse failure rates to below 5\% after exposure to 20 training examples. We show the table of the parsing failures in \autoref{section:parse_fail_rate_decrease}.

This improvement in structural consistency offers two significant advantages. First, it substantially reduces the need for granular parsing logic, simplifying the overall inference pipeline. Second, the juxtaposition of accuracy with the parse failure shows that the model gets better not only at following the prompt, but also at answering the actual test. These benefits are clear indication that fine-tuning not only improves model accuracy but also enhances output reliability for a limited number of samples.

\subsection{Impact of dataset distillation on task accuracy}
\begin{figure}[tbp]
    \centering
    \includegraphics[width=0.82\linewidth]{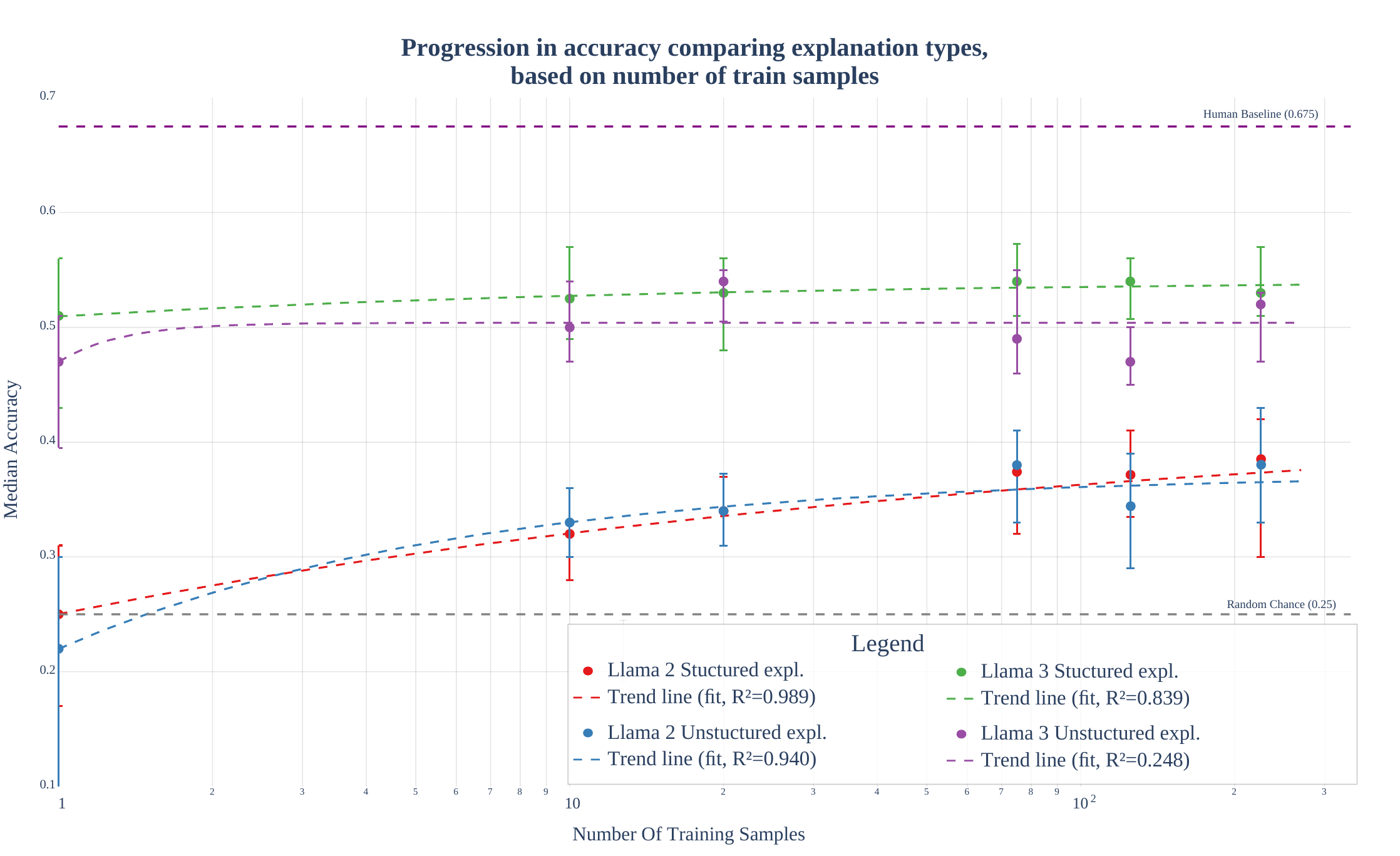}
    \caption[Learning curve of explanation distillation]{Llama 3 benefits from the structuring of the dataset whereas Llama 2 shows no clear benefit from the structuring.}
    \label{fig:learning_curve_explanation_distillation}
\end{figure}
\autoref{fig:learning_curve_explanation_distillation} shows the progressions in accuracy comparing explanation types based on number of train samples. We see marked differences between the performances of the two models for the two explanation types, and separate the performance based on the models. 

\paragraph{Impact of Structured Explanations on Llama 3} 
Llama 3 reveals markedly different learning patterns between the two explanation formats. While both approaches achieve similar accuracy ranges, the IRAC-structured format demonstrates strong correlation with training sample size ($R^2=0.839$), whereas unstructured explanations show poor correlation ($R^2=0.248$). This disparity confirms our hypothesis that the systematic IRAC structure creates more uniform training signals. The consistent trend line for structured explanations indicates that the standardized format allows the model to more reliably capture underlying patterns in legal questions as sample size increases, unlike the scattered performance observed with unstructured explanations. These results validate our distillation approach, confirming that the introduced structure improves performances, even if the content is not explicitly verified.
\paragraph{Impact of Structured Explanations on Llama 2}
Llama 2 shows minimal difference between structured and unstructured explanations throughout the training process. Both formats show strong correlation with training sample size ($R^2=0.989$ and $R^2=0.940$ respectively), indicating that while the model steadily improves with more samples, this improvement happens regardless of explanation structure. The nearly identical patterns of both approaches suggest that our distillation process, which helped Llama 3, has little impact on Llama 2's performance. This lack of difference likely comes from Llama 2's inherent limitations, including smaller context window and fewer parameters, rather than from the distillation method itself. The model seems to lack the capacity to use the structured IRAC format, limiting its ability to benefit from the more organized reasoning patterns.

\subsection{Impact of Fine-Tuning on the bias towards options}
\begin{figure}[htpb]
	\centering
	\includegraphics[width=\linewidth]{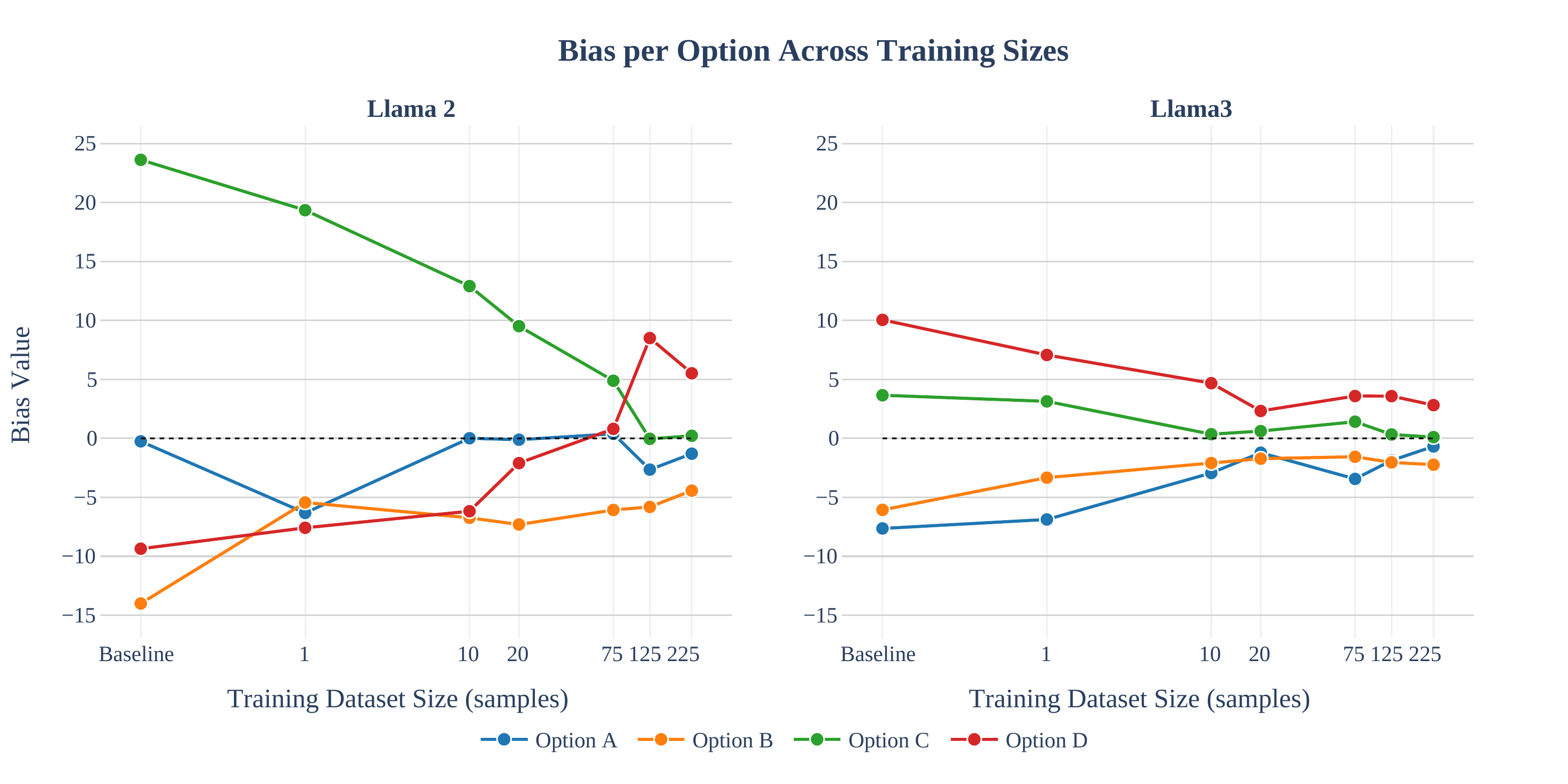}
	\caption[Progression of selection bias with fine-tuning]{As the number of added samples increases, the bias towards specific options (C for Llama 2 and D for Llama 3) decreases, indicating that we mitigate the bias as the number of training samples increases.}
	\label{fig:model_option_bias}
\end{figure}

We first try to identify if the model, out-of-the-box, shows affinity for any certain options. Once we do inference of one model instance on the test set, we then calculate the bias as the difference between the frequency of the predicted options, against the frequency of the ground truth options. This gives us a measure whether any option is chosen more than the others. For a certain SFT train sample size, we average this measure to perform analysis on the level of train samples used in SFT, if there is a decrease in certain preferences as we add more samples while fine-tuning. While this measure in and of itself is not sufficient to say if the model performs better, because it may be so that the frequency of selected options matches the frequency of the ground truth options, without answering any question correctly.
Positive bias values indicate over-selection bias and negative values indicate under-selection bias. Values closer to zero indicate that the the prediction and ground truth distributions match, but are not necessarily indicating that the model has selected the correct option.
\paragraph{Existence of inherent option bias} For the baseline, we see in \autoref{fig:model_option_bias} that the both models have preferences towards certain options, Llama 2 selecting C more often than the others, Llama 3 selecting both C and D more frequently. However, Llama 3 tends to show a smaller bias, indicating better model capability as compared to Llama 2 out of the box.
\paragraph{Reduction of bias with increasing sample size} Fine-tuning progressively reduces these inherent biases. For Llama 2, the prominent bias towards option C decreases substantially as training samples increase, reaching near-balanced levels around 75-125 samples. Similarly, Llama 3's bias towards option D diminishes with fine-tuning. This trend indicates that SFT effectively recalibrates the models' option selection tendencies even with limited samples. The matching of the distribution of predicted and ground truth options, taken alongside the accuracy improvements in \autoref{table:model-accuracy-by-samples}, indicates that the models' inherent preferences are mitigated as they actually learn the task at hand, rather than simply learning to match the statistical distribution of correct answers.

\section{Limitations and Future scope}
\paragraph{Correlation between generated explanations and the correct prediction}
While the explanations were used to guide the models towards answering the exam, we do not perform an analysis of the correlation between the correctness or factuality of the explanation and the prediction of the model. We avoid this analysis because it would either require experts to judge these questions, which would deviate from how human performance is typically evaluated on the MBE, or necessitate the use of an LLM-as-a-Judge approach. The latter introduces its own inherent biases \citep{zheng2023judgingllmasajudgemtbenchchatbot} and would require a separate verification methodology. A direct continuation of our work could analyze these explanations using multiple higher-capability models to establish the degree of correlation between the predicted option and its explanation. Furthermore, a preference dataset with labelled samples (good samples showing entailment between explanation and option, bad samples showing no entailment) could be used for further alignment using techniques like DPO to potentially improve performance.
\paragraph{Test time inference as opposed to SFT}
Since this study was confined to the fine-tuning setting, we did not investigate whether computational resources might be better invested in advanced test-time inference strategies rather than in dataset creation and model fine-tuning. The recently introduced chain-of-thought decoding \citep{wang2024chainofthoughtreasoningprompting} reports significant zero-shot performance improvements in mathematical reasoning, without requiring explicit prompting or fine-tuning. This suggests that investing additional compute during inference might yield comparable results to fine-tuning while avoiding the labor-intensive process of dataset curation. Other inference-time methods could similarly offer performance gains without the need for domain-specific training data. Future work could investigate whether these approaches provide a more efficient path to improving legal reasoning capabilities compared to supervised fine-tuning.
\paragraph{The performance of the fine-tuned models in legal situations} 
Just like in real life, mastery over the bar exam does not mean mastery over law. Thus this makes it impossible to infer that, if the model performs well in the MBE, then it can be used to support lawyers. More specific guardrails like human verification, retrieval augmented generation and other methods would be required to consider the adaptation of these models to real life, and is a non trivial task. \textit{The models we fine-tuned are only good in answering the question when framed in the format specific to the MBE}, and we have not tested for other applications, specifically chat based models, which would require a different approach to adapting these models to serve as legal assistants. 

\section{Conclusion}
In this work, we successully fine-tuned two different Llama models, to improve their performance on the Multistate Bar exam. In addition, we experimented over multiple parameters that impact the generation of the LLM, in a bid to further consolidate our results. We disilled the dataset using another more capable model, and achieved comparable performance improvements following this distillation. 
Our findings indicate a significant enhancement in model performance through supervised fine-tuning, even with minimal training data. We establish lower bounds on the number of samples needed to maximize performance for Llama 2 (7B) and Llama 3 (8B) models. While Llama 3 improves from an untrained baseline of 35.8\% to 52.5\% with just 20 samples per domain, Llama 2 improves from 18.5\% to 36.8\% but requires more samples (225) to reach its peak. Beyond accuracy gains, we observe a dramatic reduction in parsing failures with even minimal fine-tuning—from 42.7\% to 5.3\% for Llama 2 and from 30.5\% to 1.6\% for Llama 3 with just 10 samples. This improvement simplifies response extraction and eliminates the need for complex parsing strategies.

These results directly address our initial research question: our approach shows that even smaller open-weight models can achieve meaningful improvements in legal reasoning with limited datasets and consumer-grade hardware. Our comparison between Llama 2 and Llama 3 suggests that advancements in base model quality significantly impact fine-tuning outcomes, with Llama 3 showing both higher initial performance and greater sample efficiency. Though neither model matches GPT-4's performance on the MBE, our work establishes practical lower bounds for effective legal reasoning in resource-constrained environments. While larger models would likely achieve better performance, they would require additional computational resources beyond the single GPU setup used in our experiments. Our findings highlight the potential for accessible domain adaptation approaches, particularly valuable for specialized fields where collecting extensive datasets may be impractical.
\section*{Acknowledgments}
The authors acknowledge support by the state of Baden-Württemberg through bwHPC. 

\bibliography{colm2025_conference}
\bibliographystyle{colm2025_conference}
\clearpage
\appendix
\section{Detailed Analysis of GPT-4's Bar Exam Performance}
\label{app:gpt4-bar-exam}

\citet{martinez2023gpt4barperformance} highlights the following issues with the claims made in GPT-4's performance on the Uniform Bar Exam: 

\paragraph{Representation Issues in Test-Taker Comparisons}
While GPT-4's UBE score approached the 90th percentile when compared to February administrations of the Illinois Bar Exam \citep{Katz+2023}, these estimates were skewed because February test-takers typically include many repeat test-takers who failed the July administration and score significantly lower than the general test-taking population. \autoref{tab:exam-percentiles} presents a more accurate representation of GPT-4's performance across different reference populations.

\begin{table}[h]
\begin{center}
\small
\begin{tabular}{lccc}
\toprule
\multicolumn{1}{c}{\bf Test-taking population} & \multicolumn{3}{c}{\bf Corrected GPT-4 percentile on section of exam} \\
\cmidrule(lr){2-4}
& UBE & MBE & MEE + MPT \\
\midrule
July test-takers & 68th & 86th & 48th \\
All first-timers & 62rd & 79th & 42nd \\
Qualified attorneys & 45th & 69th & 15th \\
\bottomrule
\end{tabular}
\end{center}
\caption{\textbf{Percentile of the test taking population.} A more clear view emerges when looking at the performance recomputed based on the representative testing population, showing that while the performance of GPT-4 does indeed cross human thresholds, it is skewed due to the metric only being calculated for the Illinois Bar exam \citep{martinez2023gpt4barperformance}.}
\label{tab:exam-percentiles}
\end{table}

\paragraph{Evaluation Methodology for Open-Ended Responses}
The evaluation of essay examinations in \citet{Katz+2023} did not employ standardized grading rubrics typically used in actual bar examinations. Instead, comparisons were based on ``representative'' good answers published by the state of Maryland. Furthermore, while legal experts evaluated these responses, they were not specifically experienced in UBE evaluation and did not follow the blinding and calibration protocols used in official UBE grading. These methodological discrepancies, acknowledged by the authors themselves, introduce significant subjectivity into the assessment process.

\paragraph{Selective Reporting and Hyperparameter Sensitivity}
In contrast to earlier work on GPT-3.5 by \citet{bommarito2022gpt}, which investigated performance across different prompts, sampling temperatures, and token lengths, \citet{Katz+2023} presented only the best performances of GPT-4. While this approach highlighted the upper bound of the model's capabilities, it did not provide insight into performance variability across different prompting strategies, reasoning formats, or inference parameters. On the more quantifiable multiple-choice portion (MBE), \citet{martinez2023gpt4barperformance} validated GPT-4's performance at 75.66\%. \\

Another issue that we note is that LLMs present inherent interpretability challenges, as their performance is affected by multiple hyperparameters that interact in non-trivial ways. This is further complicated in closed-source models like GPT-4, where training data sources remain unverifiable. Although \citet{openai2024gpt4technicalreport} claims no contamination occurred with GPT-4's training data, this assertion cannot be independently verified, which has implications for performance evaluation on standardized tests like the bar exam.
\section{Implementation details of our approach}\label{section:approach_appendix}

\subsection{Preprocessing the data}
The datasets we gathered were separated into two documents, one containing the questions and the other containing the solutions (consisting of the correct option and the explanation for that option). We then meticulously extracted these from both documents, and strucured each question as a JSON dictionary that contained the following:
\begin{enumerate}[itemsep=1pt,parsep=0pt]
    \item The question number.
    \item Legal domain of the question.
    \item Question text body.
    \item Nested dictionary containing each option and its text.
    \item Correct option as a single string.
    \item Explanation for the solution.
\end{enumerate}
\begin{figure}[h!]
    \begin{center}
    \includegraphics[width=\linewidth]{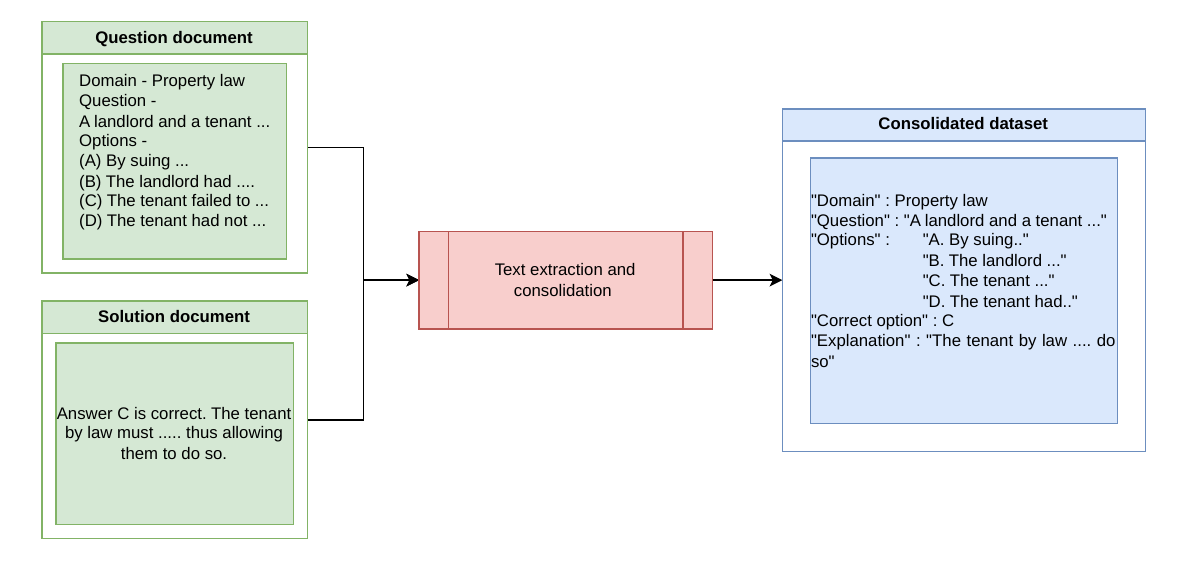}
    \end{center}
    \caption{Extraction methodology using text processing to consolidate the questions and the solutions into one structure, which we can query individual elements from. The process remained the same for both the test and train datasets.}
    \label{fig:data_extraction}
\end{figure}
Once all the questions and solutions were extracted, we were left with a JSON file that contained in each entry a dictionary with the above mentioned elements.
\subsection{The Multistate Bar exam, its contents and structure}
The MBE consists of $200$ multiple-choice questions, divided into two sessions: a morning session and an afternoon session, each lasting three hours. \emph{Only one option is considered correct}, although multiple options might be plausible, based on the difficulty of the question.
The questions cover seven areas of law, Constitutional Law, Contract Law, Criminal Law and Procedure, Evidence, Real Property, Tort Law and Civil procedure (which was included from 2015 MBE examinations onward).

\emph{We restrict our choice for the dataset to only the MBE for the sake of ease of evaluations}. The Multi-state Essay Exam (MEE) and Multi-state Performative Test (MPT) are open-ended long form writing based. The rubric for the grading of responses, while established according to standards, is dependent on human evaluations, which is not possible for us to implement owing to the challenge of scale (in terms of the number of questions an expert would have to evaluate) and the expense (in terms of number of man hours needed for evaluations).

Since the MBE has only one correct option choice for the questions, it makes it possible to evaluate our model on whether it has selected the correct option or not.
\subsection{Distribution of the questions}
The test set contains a uniform spread of questions across all the domains. However for the training set, we were able to collect relatively fewer samples for the Torts and Civil Procedure domain. For the Torts, we had originally obtained around 235 questions with solutions, but most of them were removed after de-duplication, resulting in only 69 questions. Civil procedure, on the other hand was only included as part of the MBE since 2015 examinations, and as such we were able to find only 86 questions to begin with. There were no questions from the test set that were found in the training set, and as such no questions from the train set were removed for this particular reason. 
\begin{table}[H]
\begin{center}
\begin{tabular}{lcc}
\toprule
\textbf{Legal domain} & \textbf{Train set} & \textbf{Test set} \\
\midrule
Criminal Law & 270 & 27 \\
Evidence & 272 & 30 \\
Contracts & 320 & 29 \\
Torts & 69 & 27 \\
Constitutional Law & 240 & 29 \\
Civil Procedure & 86 & 29 \\
Property & 257 & 29 \\
\midrule
\textbf{Total} & \textbf{1,514} & \textbf{200} \\
\bottomrule
\end{tabular}
\end{center}
\caption{Distribution of samples across different legal domains in the training and test sets. }\label{table:legal-domain-distribution}
\end{table}

\subsection{Modules used in the pipeline for Prompting and parsing}
We created the following modules to help with formatting the datasets into appropriate prompts based on the model, and the generation parameters. Parsing of the responses is also dependent on what prompt the model was given, and thus entails the use of a custom parsing rule for each configuration.
\subsubsection{Prompt Handler}This class deals with all things prompt related. It has been designed with the multiple possible configurations in mind and serves as a modular framework to take as input a set of configuration values and do the following 
\begin{enumerate}
    \item For SFT: Formats the training dataset into the text completion structure as is expected by the specific model. This involved the model specific special tokens and also the format that the data-sample should be in.
    \item For inference: Formats questions from the test set for the relevant experiment configuration and the model. 
\end{enumerate}
\subsubsection{Response Handler}
To simplify answer field extraction from model responses, we use regex-based text comparison rather than complex NER models. Our response handler is designed to be flexible, accommodating valid responses even when they don't strictly follow prompted structures. This approach helps avoid bias toward our fine-tuned models and maintains evaluation integrity.

The handler follows a three-step process: (1) Extract answer fields (chosen domain, option, and explanation) using primary regex patterns; (2) Apply specialized regex patterns for any extraction failures; (3) Flag cases where option or explanation fields are missing as "malformed label" or "malformed explanation" respectively. Since the chosen domain is consistently present in responses, we focus on accurately capturing the other two fields. This comprehensive approach ensures we don't discard partially valid responses when evaluating model performance.
\subsection{Further discussion about the distillation process}
\label{section:distilled_explanation}
\begin{figure}[h!]
	\centering
	\includegraphics[width=0.8\linewidth]{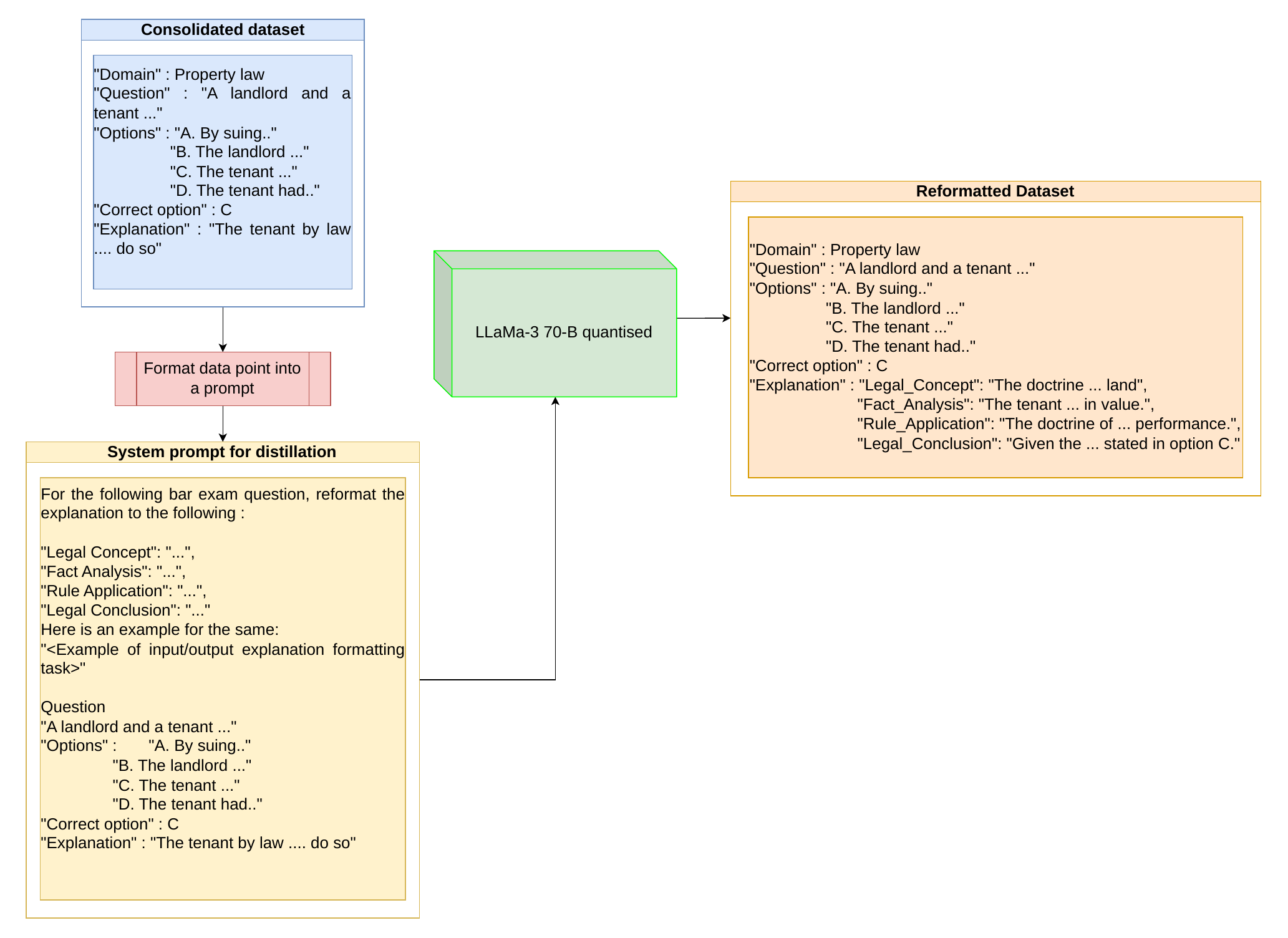}
	\caption[Data distillation process]{Data distillation done using Llama 3 70B to restructure the explanation into IRAC format.}
	\label{fig:data_distillation}
\end{figure}

\paragraph{Veracity of the distilled explanation}We do not verify the correctness of the restructured explanation. It might be entirely possible that the distilling model might introduce its own hallucinations. We expect the provided context from the explanation, along with the solution and the questions, to be sufficient in guiding the model to perform the restructuring. In any case, whether the generated explanation is wholly correct or not, our main goal with this restructuring is to introduce the fine-tuned models to this pattern, hopefully improving their performance against the models fine-tuned with the unstructured explanation.
In the example \autoref{fig:textbox-comparison} from the train set, the 'Dormant Commerce Clause', which is not present in the original explanation has been introduced. This is a direct addition from the model without any further prompting. The few shot example to the model, was crafted with the use of GPT-4, with Claude (both Sonet and Haiku versions) being used as referees for qualitative analysis, and only the crafted example that had consensus among the two models was used.\\
\autoref{fig:data_distillation} shows our approach and how the dataset was prompt formatted with a few shot example showing the task at hand.
\begin{figure}[H]
	\centering
	\fbox{%
		\begin{minipage}{1.0\textwidth}
			\textbf{Raw explanation:} \\
			States may not impose restrictions that impede the flow of interstate commerce by enacting legislation that protects its institutions or residents from competition, which is what this statute does. B is incorrect because the requirement that candidates be graduates of in-state colleges is not a classification based on residency. Both in- and out-of-state residents must be graduates of the state colleges.\\
			
			\vspace{1em} 
			\hrule 
			\vspace{1em} 
			
			\textbf{Processed explanation}: \\
			\textit{\textbf{Legal Concept}}: Dormant Commerce Clause and the principle of interstate commerce.\\ 
			\textit{\textbf{Fact Analysis}}: The state has implemented a licensure requirement for barbers, mandating that they must graduate from barber schools located within the state, with the goal of ensuring quality standards. \\
			\textit{\textbf{Rule Application}}: The Dormant Commerce Clause prohibits states from enacting laws that unduly burden or discriminate against interstate commerce. In this case, the requirement favors in-state barber schools over out-of-state schools, potentially hindering the flow of commerce and competition. \\
			\textit{\textbf{Legal Conclusion}}: The requirement that barber licenses be restricted to graduates of in-state barber schools is likely unconstitutional as an undue burden on interstate commerce, supporting option A. 
			
		\end{minipage}}
	\caption[Comparison of explanations before and after formatting]{Comparison of unstructured and structured explanation obtained after distillation}
	\label{fig:textbox-comparison}
\end{figure}

\subsection{Fine-tuning Hyperparameters and training times}\label{section:hyperparams}
\autoref{tab:qft_hyperparams} shows the fine-tuning specific hyper parameters. For the quantised lora, we load the model in 'NF4' quantisation in order to load more batches with the saved memory. The total memory load on the GPU for the fine-tuning reaches a maximum of 28.8 GB for Llama 3 and 19GB for Llama 2. Our choice of using a Cosine Scheduler with warm restarts for the learning rate instead of a constant learning rate stems from exploratory runs, where constant LR didn't show promising results. We did not do a sweep over different learning rates and relegate it to future works to find the best hyperparameter settings for fine-tuning. 
While running inference, we load the models in full precision, with Llama 3 taking up 30GB and Llama 2 taking up 27 GB memory. 

\begin{table}[h!]
\centering
\renewcommand{\arraystretch}{1.5}
\setlength{\tabcolsep}{12pt}
\begin{tabular}{lll}
\hline
\textbf{Category} & \textbf{Hyperparameter} & \textbf{Value} \\
\hline
\multirow{6}{*}{LoRA Config} 
    & Rank ($r$) & 64 \\
    & Alpha & 32 \\
    & Bias & None \\
    & Dropout & 0.05 \\
    & Task Type & Causal LM \\
    & Target Modules & \begin{tabular}[c]{@{}l@{}}q\_proj, k\_proj, v\_proj, o\_proj,\\
                                          up\_proj, down\_proj, gate\_proj\end{tabular} \\
\hline
\multirow{9}{*}{Training Args} 
    & Epochs & 10 \\
    & Batch Size & 7 \\
    & Gradient Accumulation & 1 \\
    & Learning Rate & 0.0001 \\
    & Weight Decay & 0.01 \\
    & Max Gradient Norm & 0.3 \\
    & LR Scheduler & Cosine with Restarts \\
    & Warmup Ratio & 0.1 \\
    & Cycles & 2 \\
\hline
\end{tabular}
\caption{QLoRA Training Hyperparameters}
\label{tab:qft_hyperparams}
\end{table}
\autoref{table:training_times} show the approximate time taken per SFT run, with the SFT and PEFT hyperparameters remaining the same for every run. 
\begin{table}[H]
\begin{center}
\begin{tabular}{lc}
\toprule
\textbf{Train Sample Size} & \textbf{Average Training time (HH:MM:SS)} \\
\midrule
1                 & 00:06:00                          \\
10                & 00:40:00                          \\
20                & 02:15:00                          \\
75                & 07:15:00                          \\
125               & 11:30:00                          \\
225               & 21:40:00                          \\
\bottomrule
\end{tabular}
\end{center}
\caption{Approximate fine-tuning times for different train sample sizes.}
\label{table:training_times}
\end{table}

\newpage
\section{Further Results}

\subsection{Performance of the untrained baseline models} 
\subsubsection{How often are the labels not parseable for the baseline models?}
\begin{figure}[h!]
    \centering
    \includegraphics[width=1.1\linewidth]{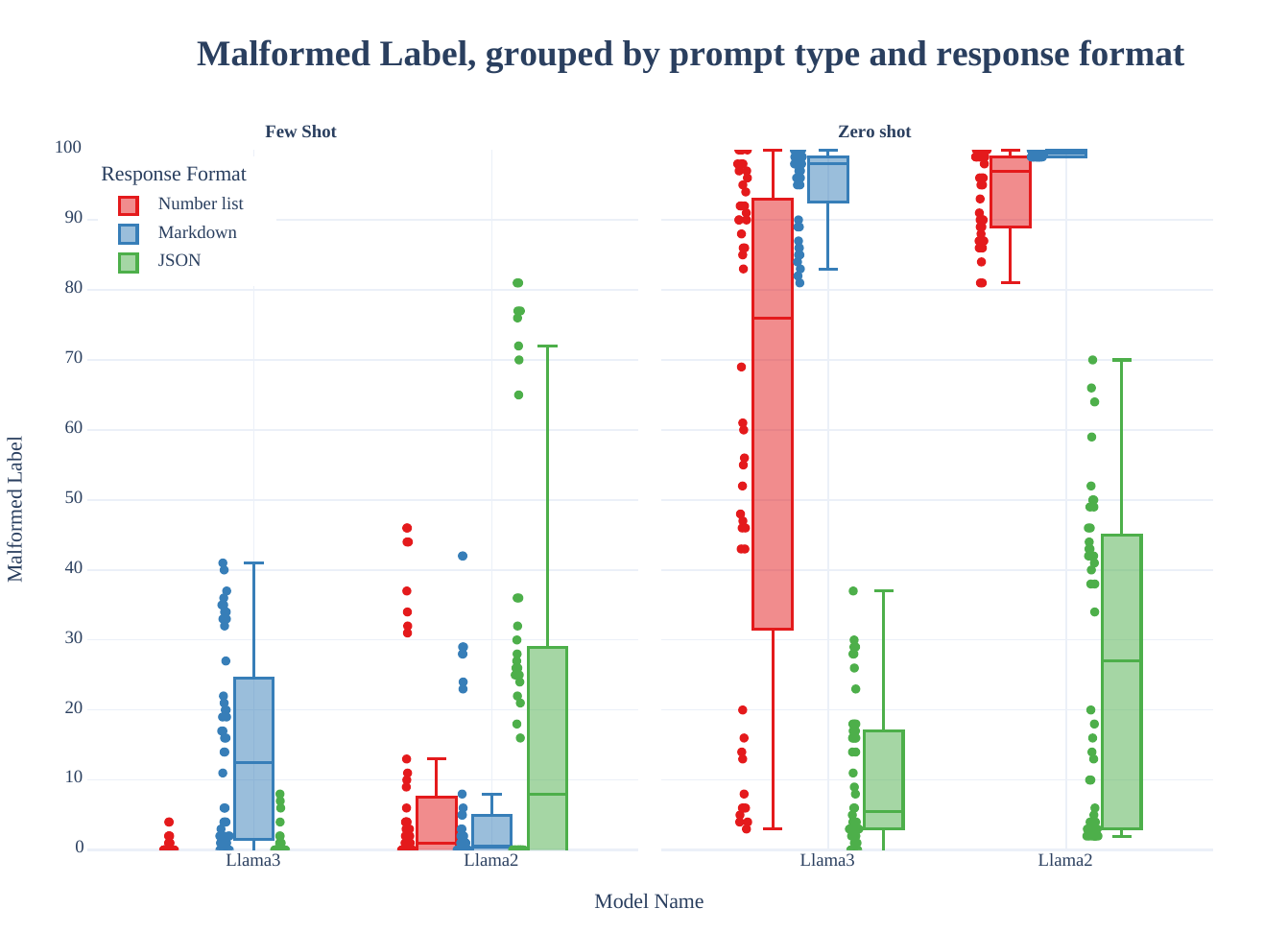}
    \caption[Parsing failures in untrained baselines]{Llama 3 has significantly fewer parsing failures as compared to Llama 2, with both models having lesser parsing failures in general with a few-shot prompt.}
    \label{fig:untrained_malformed_label}
\end{figure}

The secondary issue that the models must overcome, is their adherence to the prompt. We consider a response in which the selected label cannot be extracted as a wrong answer, given the reasoning that the response parser is made to be as relaxed as possible, and some responses simply do not contain the label, as opposed to other responses where the predicted option label exists but doesn't conform to the expected structure. Since we already handle the latter case, we can still have reliable comparison where the model is afforded some flexibility in generation for the untrained case. We see in \autoref{fig:untrained_malformed_label}, that for both Llama 2 and Llama 3, the number of parsing failures is higher for the zero-shot prompt case. In the few-shot prompt, when an example of the task is provided, both models have a significantly lower parse failure rate, showing the importance of providing the example. Curiously, for both the models, JSON elicits lesser parsing failures in the zero shot case, as compared to its counterparts. This is mostly due to the inherent structural constraints of the JSON structure, which narrows done slightly the possible tokens to generate sequentially, thus allowing for the response to be more closed as compared to the other two formats.\\
\autoref{table:model-performance-table} shows a side-by-side comparison of the correct predictions and the malformed labels, showing that the accuracy of both models is hurt by the number of parsing failures, and they have a large variance. While we can control for the structure of the response format, we do not try out different wordings and structures in the prompts to find the most optimal prompt, as that would require an effort of its own.  
\begin{table}[h!]
\begin{center}
\begin{tabular}{lcccccccccc}
\toprule
\multirow{2}{*}{\textbf{Model}} & \multicolumn{5}{c}{\textbf{Malformed Label}} & \multicolumn{5}{c}{\textbf{Correct Predictions}} \\
\cmidrule(lr){2-6} \cmidrule(lr){7-11}
& \textbf{Mean} & \textbf{Std} & \textbf{Median} & \textbf{Min} & \textbf{Max} & \textbf{Mean} & \textbf{Std} & \textbf{Median} & \textbf{Min} & \textbf{Max} \\
\midrule
Llama 2 & 42.72 & 42.32 & 28.50 & 0 & 100 & 18.50 & 14.37 & 22 & 0 & 42 \\
Llama 3 & 30.47 & 39.19 & 6.00 & 0 & 100 & 35.66 & 20.49 & 45 & 0 & 61 \\
\bottomrule
\end{tabular}
\end{center}
\caption{Performance comparison of number of correct predictions and failed parsed responses among the baseline models. The more recent Llama 3 model has lesser parsing issues and higher accuracies out of the box.}\label{table:model-performance-table}
\end{table}
\subsection{Decrease in parsing failures with increasing sample sizes}\label{section:parse_fail_rate_decrease}
We see in \autoref{table:malformed-labels-by-samples} how as fine-tuning incorporates more train samples, we achieve a sharp decrease in the frequency of the parsing failures from the model responses for both the models. This decrease in allows our accuracy estimates be more representative of the models' reasoning on the MBE. We see that there is a stark reduction with even 1-sample SFT. Llama 3 is better able to adjust to the fine-tuning, whereas Llama 2 still shows some variance and parsing failures are still an issue, albeit not as significant, for larger sample training regimes. 
\begin{table}[!h]
\begin{center}
\begin{tabular}{lcccccccc}
\toprule
\multirow{2}{*}{\textbf{Samples}} & \multicolumn{4}{c}{\textbf{Llama 2}} & \multicolumn{4}{c}{\textbf{Llama 3}} \\
\cmidrule(lr){2-5} \cmidrule(lr){6-9}
& \textbf{Mean} & \textbf{Std} & \textbf{Median} & \textbf{Max} & \textbf{Mean} & \textbf{Std} & \textbf{Median} & \textbf{Max} \\
\midrule
Untrained baseline & 42.72 & 42.32 & 28.5 & 100 & 30.47 & 39.19 & 6.0 & 100 \\
1 & 30.98 & 30.79 & 21.0 & 89 & 12.50 & 17.98 & 4.0 & 79 \\
10 & 5.34 & 14.95 & 0.0 & 80 & 1.60 & 4.10 & 0.0 & 24 \\
20 & 2.47 & 7.43 & 0.0 & 49 & 1.60 & 4.03 & 0.0 & 25 \\
75 & 2.43 & 4.41 & 1.0 & 21 & 0.56 & 1.11 & 0.0 & 7 \\
125 & 2.48 & 3.54 & 0.0 & 14 & 0.74 & 1.27 & 0.0 & 5 \\
225 & 4.99 & 8.38 & 0.0 & 35 & 0.73 & 1.31 & 0.0 & 7 \\
\bottomrule
\end{tabular}
\end{center}
\caption{Comparison of malformed label occurrences across different training sample sizes. Lower values indicate better parsing performance. }\label{table:malformed-labels-by-samples}
\end{table}
\subsection{How do the generation parameters affect the performance of the fine-tuned models?}\label{section:parameter-effects}
\subsubsection{The effect of different temperatures on the performance}\label{section:temperature_effect}
\begin{figure}[h!]
	\centering
	\includegraphics[width=\linewidth]{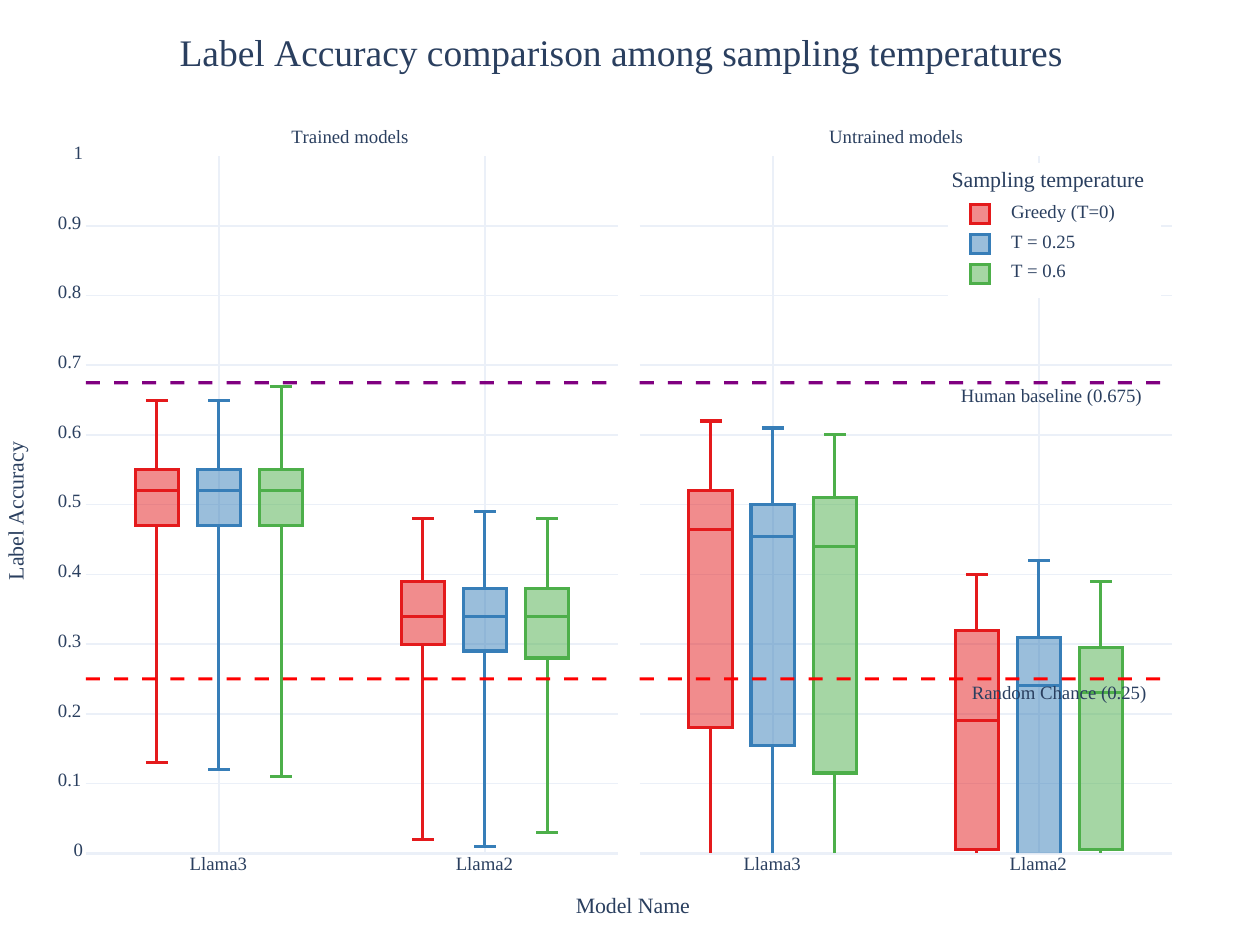}
	\caption[Comparing Accuracy against different temperatures]{There is an insignificant difference between the sampling temperatures showing that the temperature used in temperature based sampling doesn't affect much the performance on the task.}
	\label{fig:accuracy_by_temperature}
\end{figure}

While the temperature that affects the generation in the default top-p sampling was initially a hyperparameter to test its effect on the accuracy, we see in \autoref{fig:accuracy_by_temperature} that there is no significant difference in accuracy. Thus, while considering the accuracy, we average the metric  over these parameters, along with the two seeds over which we ran inference. 
\subsubsection{Qualitative analysis of best performing parameter configuration types across different train sample sizes}
\paragraph{Llama 3's best configurations} We see the following counts in \autoref{table:Llama 3_best_configs}:
\begin{enumerate}
    \item Prompt type: \textbf{Few-shot (15)} vs Zero-shot (6)
    \item Explanation type: \textbf{Structured (16)} vs Unstructured (5)
    \item Response format: \textbf{JSON (9)} vs Markdown (6) vs Numbered List (6)
    \item Response type: Answer first (8) vs \textbf{Fact first (13)} 
\end{enumerate}
While the complex interactions of factors other than the parameters we considered are difficult to account for, the frequency of each of these parameters in the best performing configurations allows for a simpler comparison, showing that in alignment with our hypothesis, the structuring of the explanation that we obtain from the distillation does in fact benefit the model, giving it a better grasp over the task and leading to consistent higher performances. The Prompt type also does seem to help the model in lower data regimes, but with progressive training we see more zero shot prompt configurations perform better over few shot configurations, which is indicative of the model depending lesser on the quality of the prompt and more on the fine-tuning process. Less significant is the response format, with a slightly higher frequency of JSON structured response format instances performing slightly better, as compared to markdown or numbered list. This however, bodes well in terms of the complexity of parsing responses, as existing JSON parsing libraries can effortlessly parse a correct JSON structured response, as opposed to using regular expressions for the other two response formats. Another interesting thing is that the Fact-first response type yields higher ranking performances, indicating that when the model first provides an explanation, it yields higher performance on average. We see a higher pairing of fact first, structured type models, indicating that the elucidation of the explanation in a distilled format, before the chosen option yields much better performance, showing that Llama 3 benefits from this, possibly due to its better pretraining and larger context length. 

\begin{table}[H]
\renewcommand{\arraystretch}{1.3} 
\setlength{\tabcolsep}{4pt} 
\begin{tabular}{ccccccc}
\hline
\multirow{2}{*}{\textbf{Train sample}} & \multirow{2}{*}{\textbf{Rank}} & \multicolumn{4}{c}{\textbf{Parameters}} & \multirow{2}{*}{\shortstack{\textbf{Best} \\ \textbf{Avg. Acc.}}} \\ \cline{3-6}
                                       &                              & \textbf{Prompt type} & \textbf{Explanation type} & \textbf{Response type} & \textbf{Response format} &                                     \\ \hline
\multirow{3}{*}{Baseline}              & 1                            & Few shot              & Unstructured               & Fact first              & Json                       & 56.88                               \\
                                       & 2                            & Few shot              & Structured                 & Fact first              & Number list                & 54.50                               \\
                                       & 3                            & Few shot              & Unstructured               & Fact first              & Markdown                   & 54.50                               \\ \hline
\multirow{3}{*}{1 sample}               & 1                            & Few shot              & Structured                 & Fact first              & Json                       & 56.50                               \\
                                       & 2                            & Few shot              & Structured                 & Fact first              & Markdown                   & 56.25                               \\
                                       & 3                            & Few shot              & Structured                 & Fact first              & Number list                & 56.00                               \\ \hline
\multirow{3}{*}{10 samples}             & 1                            & Few shot              & Structured                 & Answer first            & Number list                & 57.25                               \\
                                       & 2                            & Few shot              & Structured                 & Fact first              & Markdown                   & 57.08                               \\
                                       & 3                            & Few shot              & Structured                 & Fact first              & Json                       & 55.75                               \\ \hline
\multirow{3}{*}{20 samples}             & 1                            & Few shot              & Structured                 & Fact first              & Json                       & 59.00                               \\
                                       & 2                            & Few shot              & Unstructured               & Answer first            & Json                       & 54.90                               \\
                                       & 3                            & Zero shot             & Unstructured               & Fact first              & Number list                & 54.83                               \\ \hline
\multirow{3}{*}{75 samples}             & 1                            & Few shot              & Structured                 & Answer first            & Markdown                   & 58.50                               \\
                                       & 2                            & Few shot              & Unstructured               & Answer first            & Markdown                   & 56.75                               \\
                                       & 3                            & Zero shot             & Structured                 & Answer first            & Json                       & 56.16                               \\ \hline
\multirow{3}{*}{125 samples}            & 1                            & Zero shot             & Structured                 & Fact first              & Json                       & 55.75                               \\
                                       & 2                            & Zero shot             & Structured                 & Fact first              & Number list                & 55.25                               \\
                                       & 3                            & Few shot              & Structured                 & Answer first            & Json                       & 55.16                               \\ \hline
\multirow{3}{*}{All samples}            & 1                            & Zero shot             & Structured                 & Fact first              & Markdown                   & 55.66                               \\
                                       & 2                            & Zero shot             & Structured                 & Answer first            & Json                       & 55.41                               \\
                                       & 3                            & Few shot              & Structured                 & Answer first            & Number list                & 55.41                               \\ \hline
\end{tabular}
\caption{Best performing configurations for Llama 3 and their accuracy, averaged over the seed and the temperature.}
\label{table:Llama 3_best_configs}
\end{table}
\paragraph{Llama 2's best configurations} We see the following in \autoref{table:Llama 2_best_performing_configs}:
\begin{enumerate}
    \item Prompt type: Few-shot (10) vs \textbf{Zero-shot (11)}
    \item Explanation type: \textbf{Structured (11)} vs Unstructured (10)
    \item Response format: \textbf{JSON (10)} vs Markdown (4) vs Numbered List (7)
    \item Response type: \textbf{Answer first (15)} vs Fact first (6) 
\end{enumerate}
To break it down, we see that zero shot prompting is more frequently ranked higher at bigger train sample sizes, just like for Llama 3 is more frequently better at lower train sample regimes, indicating the same dependence of the model on the prompt quality (consisting of the example task) to help the accuracy, which indicates lesser dependence on the prompt as training sample size increases. The possible explanation as to why the few shot prompting doesnt help, could be that the extra tokens taken up by the example offered in theprompt must be affecting the model attention adversely, as it has a smaller context length of 2048 tokens as compared to the 4096 tokens that Llama 3 is capable of. We also see, that the model does not seem to gain an edge using structured explanations, and the rate of wins is similar to when unstructured explanations are used. This could possibly be because of the smaller context length and overall lower capability of the model against the Llama 3 model. It shows that these parameters dont necessarily provide the Llama 2 model any edges and better performance can only possibly be obtained by increasing the number of samples used in the training, which is evident from the monotonous increase in the accuracy. 
\begin{table}[h!]
\renewcommand{\arraystretch}{1.3} 
\setlength{\tabcolsep}{4pt} 
\begin{tabular}{ccccccc}
\hline
\multirow{2}{*}{\textbf{Train sample}} & \multirow{2}{*}{\textbf{Rank}} & \multicolumn{4}{c}{\textbf{Parameters}} & \multirow{2}{*}{\shortstack{\textbf{Best} \\ \textbf{Avg. Acc.}}} \\ \cline{3-6}
                                       &                              & \textbf{Prompt type} & \textbf{Explanation type} & \textbf{Response type} & \textbf{Response format} &                                     \\ \hline
\multirow{3}{*}{Baseline}              & 1                            & Few shot              & Structured                 & Fact first              & Number list                & 33.83                               \\
                                       & 2                            & Few shot              & Structured                 & Answer first            & Json                       & 32.91                               \\
                                       & 3                            & Few shot              & Unstructured               & Answer first            & Json                       & 32.33                               \\ \hline
\multirow{3}{*}{1 sample}               & 1                            & Few shot              & Unstructured               & Answer first            & Json                       & 35.55                               \\
                                       & 2                            & Few shot              & Unstructured               & Answer first            & Markdown                   & 33.50                               \\
                                       & 3                            & Few shot              & Structured                 & Fact first              & Number list                & 31.66                               \\ \hline
\multirow{3}{*}{10 samples}             & 1                            & Zero shot             & Structured                 & Fact first              & Json                       & 36.75                               \\
                                       & 2                            & Zero shot             & Structured                 & Fact first              & Number list                & 35.75                               \\
                                       & 3                            & Few shot              & Unstructured               & Answer first            & Markdown                   & 35.58                               \\ \hline
\multirow{3}{*}{20 samples}             & 1                            & Zero shot             & Unstructured               & Answer first            & Json                       & 39.00                               \\
                                       & 2                            & Few shot              & Structured                 & Fact first              & Number list                & 38.36                               \\
                                       & 3                            & Zero shot             & Structured                 & Fact first              & Markdown                   & 38.08                               \\ \hline
\multirow{3}{*}{75 samples}             & 1                            & Zero shot             & Unstructured               & Answer first            & Json                       & 44.75                               \\
                                       & 2                            & Zero shot             & Unstructured               & Answer first            & Number list                & 43.33                               \\
                                       & 3                            & Few shot              & Unstructured               & Answer first            & Json                       & 40.66                               \\ \hline
\multirow{3}{*}{125 samples}            & 1                            & Zero shot             & Structured                 & Answer first            & Number list                & 41.75                               \\
                                       & 2                            & Zero shot             & Structured                 & Answer first            & Json                       & 41.00                               \\
                                       & 3                            & Few shot              & Structured                 & Answer first            & Markdown                   & 40.58                               \\ \hline
\multirow{3}{*}{All samples}            & 1                            & Zero shot             & Unstructured               & Answer first            & Json/Number list           & 45.08                               \\
                                       & 2                            & Zero shot             & Unstructured               & Answer first            & Markdown                   & 43.58                               \\
                                       & 3                            & Zero shot             & Structured                 & Answer first            & Json                       & 42.88                               \\ \hline
\end{tabular}
\caption{Best performing configurations for Llama 2 and their accuracy, averaged over the seed and the temperature.}
\label{table:Llama 2_best_performing_configs}
\end{table}
\subsection{Overall RMS bias as training progresses}
\begin{figure}[H]
	\centering
	\includegraphics[width=0.68\linewidth]{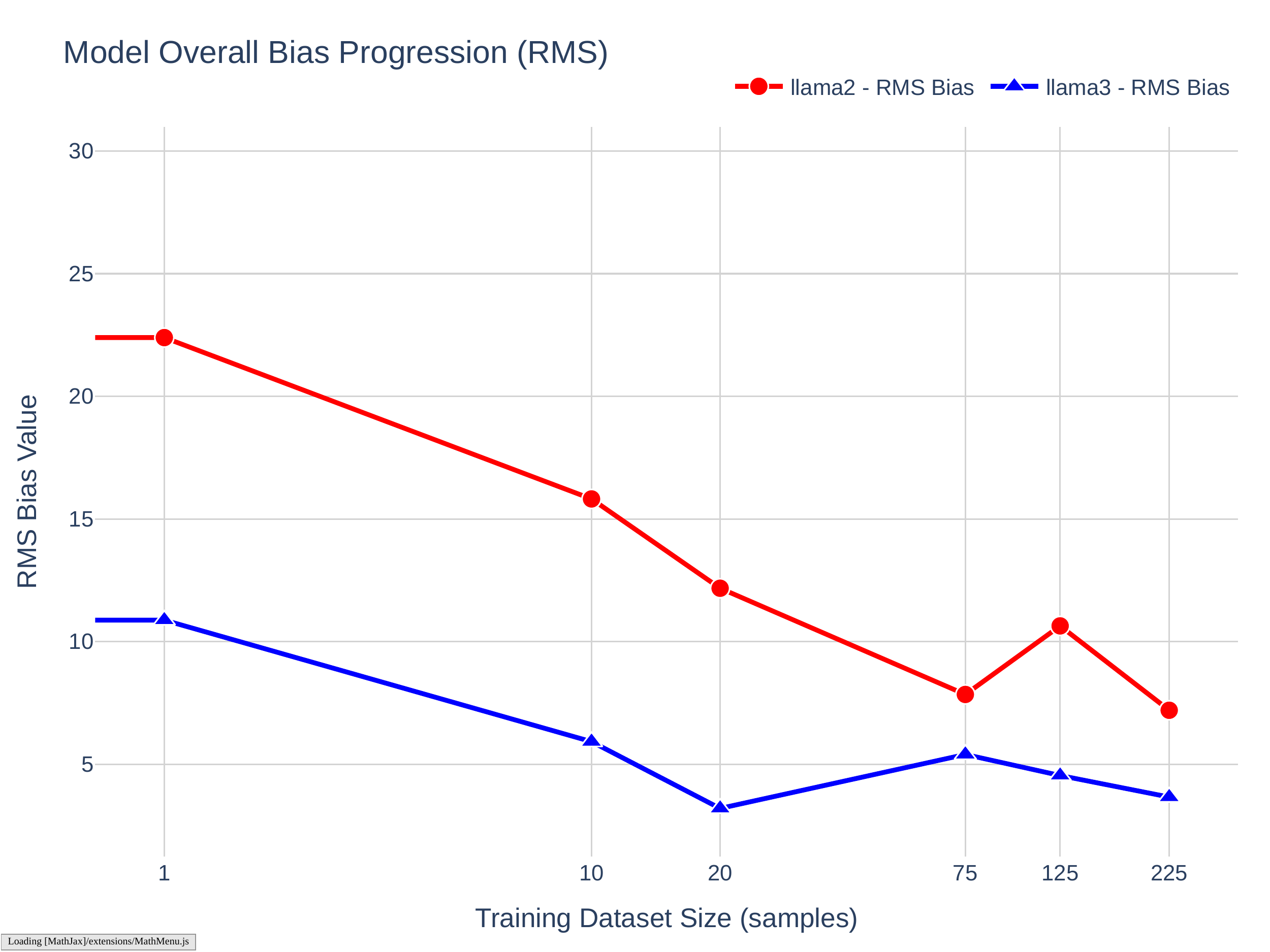}
	\caption[Overall Bias decrease with increase sample size]{Overall Bias decrease with increase sample size}
	\label{fig:model_rms_bias_progression}
\end{figure}

We  plot the overall bias as 
\begin{equation}
    \text{Bias}^{}_{RMS} = \frac{1}{4}\sqrt{\sum_{j \in \{A,B,C,D\}} \text{Bias}_j^2}
\end{equation}
We see in \autoref{fig:model_rms_bias_progression} that for the number of training samples, while there is a general decrease in the overall bias, the least bias for both the models is reported for the number of samples that yield the best performance (20 samples for Llama 3 and 225 samples for Llama 2), indicating that even though the model may select the wrong option, it is not doing so based on inherent preferences towards certain options. 
\subsection{Domain based analysis of model performance}
Since we do not have the baselines according to human test takers based on the legal domain, we consider the performance on the bar in terms of the improvement that the models gain over the baselines with increasing sample sizes. Our results in \autoref{table:domain-based-accuracy-llama-comparison} clearly show that fine-tuning provides an immediate boost to the performance on each domain, however we see that the increase in performance across each domain cannot be tied down to any one training sample size . Since we do not try to interpret the questions from our training or test sets, we cannot interpret the performances of the models in terms of the difficulty of each domain and hence present this information as is. 

\begin{table}[H]
\begin{center}
\begin{tabular}{lccccccc}
\toprule
\textbf{Samples} & \textbf{Civil} & \textbf{Constitutional} & \textbf{Contracts} & \textbf{Criminal} & \textbf{Evidence} & \textbf{Real} & \textbf{Tort} \\
\midrule
\multicolumn{8}{c}{\textbf{Llama 3}} \\
\midrule
0 & 8.91 & 10.24 & 5.85 & 13.17 & 11.45 & 11.43 & 14.68 \\
1 & 16.25 & 31.31 & 16.41 & 27.93 & 22.38 & 19.72 & 34.33 \\
10 & 39.47 & \cellcolor{blue!25}56.83 & 34.19 & 52.15 & 43.39 & 39.52 & \cellcolor{lightyellow}55.11 \\
20 & \cellcolor{lightyellow}43.16 & \cellcolor{lightyellow}55.61 & 34.76 & 55.15 & \cellcolor{blue!25}49.74 & 43.88 & 54.02 \\
75 & \cellcolor{blue!25}46.52 & 51.10 & \cellcolor{lightyellow}37.28 & \cellcolor{lightyellow}59.62 & 47.31 & \cellcolor{lightyellow}47.57 & \cellcolor{blue!25}56.80 \\
125 & 42.28 & 47.58 & 36.45 & 52.64 & 46.81 & 44.70 & 54.95 \\
225 & 45.61 & 47.95 & \cellcolor{blue!25}39.18 & \cellcolor{blue!25}60.42 & \cellcolor{lightyellow}49.47 & \cellcolor{blue!25}50.62 & 50.57 \\
\midrule
\midrule
\multicolumn{8}{c}{\textbf{Llama 2}} \\
\midrule
0 & 8.33 & 10.39 & 14.05 & 15.94 & 4.18 & 4.35 & 9.57 \\
1 & 7.84 & 12.62 & 15.08 & 18.09 & 5.18 & 7.29 & 11.15 \\
10 & 26.71 & 28.99 & 33.48 & 33.79 & 21.13 & 25.25 & 26.68 \\
20 & \cellcolor{blue!25}29.08 &  \cellcolor{lightyellow}30.41 & \cellcolor{lightyellow}37.75 & 36.68 & 23.89 & 28.45 & 29.22 \\
75 & \cellcolor{lightyellow}27.37 & \cellcolor{blue!25}31.76 & \cellcolor{blue!25}39.61 & \cellcolor{lightyellow}38.62 & \cellcolor{blue!25}27.54 & \cellcolor{lightyellow}32.74 & 29.32 \\
125 & 22.71 & 28.47 & 37.57 & \cellcolor{blue!25}39.14 & \cellcolor{lightyellow}27.00 & 31.77 & \cellcolor{blue!25}30.29 \\
225 & 22.64 & 30.02 & 35.81 & 36.80 & 26.10 & \cellcolor{blue!25}33.52 & \cellcolor{lightyellow}30.05 \\
\bottomrule
\end{tabular}
\end{center}
\caption{Accuracies per domain for Llama 3 and Llama 2 across different numbers of training samples. The highest/2nd-highest accuracy for each domain is highlighted in \colorbox{blue!25}{blue}/\colorbox{lightyellow}{yellow}.}
\label{table:domain-based-accuracy-llama-comparison}
\end{table}

\newpage
\section{System prompts for different configurations}
The following shows the basic system prompt for the configurations based on the response format ( JSON, Markdown or Numbered list) for different explanation structures. Based on this, if there is an example to be provided, it is add to the system prompt based on the experimental configuration that the prompt is made for. We show here the fact first configuration, and the answer first is simply a reordering of the chosen option to be generated before the explanation. 
\begin{tcolorbox}[colback=white!95!black, colframe=black, title= JSON | Fact-First | Unstructured explanation, sharp corners, boxrule=0.5mm, width=\textwidth]
	\begin{lstlisting}[breaklines=true]
Respond to the Multi state Bar exam question into this STRICT JSON format, considering the question and answer choices provided:

{
	"chosen_domain": "*The relevant legal domain for this question*",
	"explanation": "*Analysis of the facts supporting the best possible answer choice for the scenario, taking into account relevant laws*"
	"chosen_option_label": "*The letter corresponding to the correct answer choice.*"
}
	\end{lstlisting}
	
\end{tcolorbox}
\begin{tcolorbox}[colback=white!95!black, colframe=black, title=JSON | Fact-First | Structured explanation, sharp corners, boxrule=0.5mm, width=\textwidth][H]
	\begin{lstlisting}[breaklines=true]
Respond to the Multi state Bar exam question into this STRICT JSON format, considering the question and answer choices provided:
	
{
"chosen_domain": "The relevant legal domain for this question",
"explanation": {
	"Legal_Concept": "*The legal principle or doctrine at the heart of the question.*",
	"Fact_Analysis": "*A breakdown of the relevant facts presented in the question*",
	"Rule_Application": "*How the legal concept applies to thefacts in the question.*",
	"Legal_Conclusion": "*A concise statement explaining why the chosen option is the correct answer.*"},
"chosen_option_label": "*The letter corresponding to the correct answer choice.*"
}
	\end{lstlisting}
	
\end{tcolorbox}

\begin{tcolorbox}[colback=white!95!black, colframe=black, title= Markdown | Fact-First | Structured explanation, sharp corners, boxrule=0.5mm, width=\textwidth]
	\begin{lstlisting}[breaklines=true]
Respond to the Multistate Bar Exam question with a detailed explanation in the following MARKDOWN format, considering the question and answer choices provided:

## Domain
*The relevant legal domain for this question.*

## Explanation

### Legal Concept
*The legal principle or doctrine at the heart of the question.*

### Fact Analysis
*A breakdown of the relevant facts presented in the question.*

### Rule Application
*How the legal concept applies to the facts in the question.*

### Legal Conclusion
*A concise statement explaining why the chosen option is the correct answer.*

## Chosen Option
*The letter corresponding to the correct answer choice.* 

	\end{lstlisting}
	
\end{tcolorbox}

\begin{tcolorbox}[colback=white!95!black, colframe=black, title= Numbered list | Fact-First | Structured explanation, sharp corners, boxrule=0.5mm, width=\textwidth]
	\begin{lstlisting}[breaklines=true]
Answer the following Multistate Bar Exam question. Provide the following information in your response:

1. Chosen Domain: (The relevant legal domain for this question) 
2. Legal Concept: (The legal principle or doctrine at the heart of the question)
3. Fact Analysis: (A breakdown of the relevant facts presented in the question)
4. Rule Application: (How the legal concept applies to the facts in the question)
5. Legal Conclusion: (A concise statement explaining why the chosen option is the correct answer)
6. Chosen Option Label: (The letter corresponding to the correct answer choice)
	\end{lstlisting}
\end{tcolorbox}	
\begin{tcolorbox}[colback=white!95!black, colframe=black, title= Markdown | Fact-First | Unstructured explanation, sharp corners, boxrule=0.5mm, width=\textwidth]
	\begin{lstlisting}[breaklines=true]
Respond to the Multistate Bar Exam question with a detailed explanation in the following MARKDOWN format, considering the question and answer choices provided:

## Domain
*The relevant legal domain for this question.*

## Explanation
*Analysis of the facts supporting the best possible answer choice for the scenario, taking into account relevant laws*

## Chosen Option
*The letter corresponding to the correct answer choice.* 


	\end{lstlisting}
	
\end{tcolorbox}

\begin{tcolorbox}[colback=white!95!black, colframe=black, title= Numbered list | Fact-First | Unstructured explanation, sharp corners, boxrule=0.5mm, width=\textwidth]
	\begin{lstlisting}[breaklines=true]
Answer the following Multistate Bar Exam question. Provide the following information in your response:

1. Chosen Domain: (The relevant legal domain for this question) 
2. Explanation: (Analysis of the facts supporting the best possible answer choice for the scenario, taking into account relevant laws and principles)
3. Chosen Option Label: (The letter corresponding to the correct answer choice)
	\end{lstlisting}
	
\end{tcolorbox}

\section{Hallucinations and other errors in responses}
We identify the most common hallucinations and other errors that result in the baselines, which both the models share. Some of these hallucinations are benign in the sense that they are easily sidestepped by our response handler while extracting the responses, but some of them effectively block the extraction and hence lead to the errors that we see in parsing. 
\subsection{Sentence repetition and inability to terminate generation.}
\begin{figure}[H]
	\centering
	\includegraphics[width=1.0\linewidth]{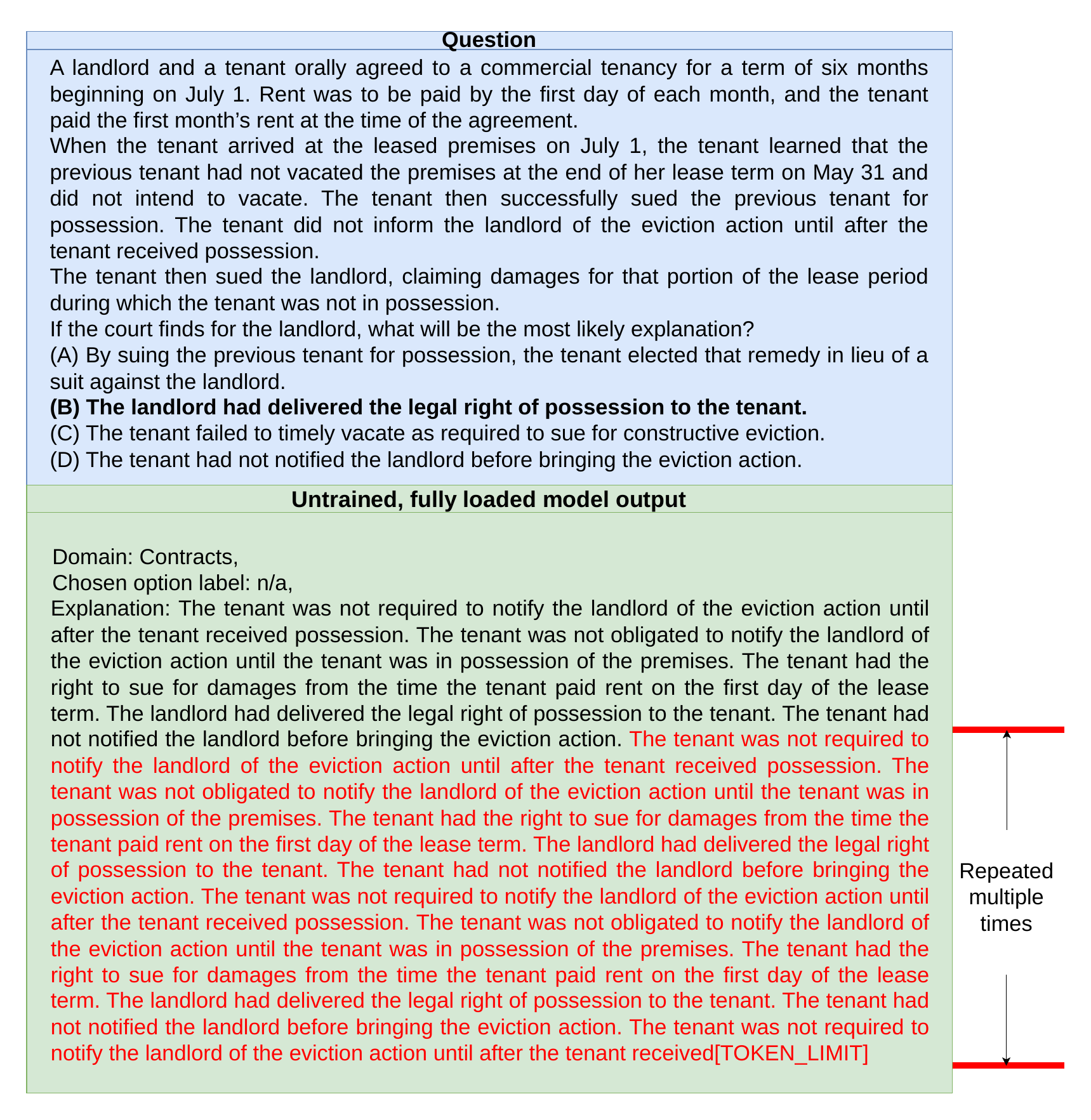}
	\caption[Wrongly parsed model output for baseline model]{This is the extracted, parsed output. The model does not know when to stop, and keeps recursing over the same tokens, thus resulting in the option label not being generated as this specific scenario is a 'Fact-first' configuration.	}
	\label{fig:untrained_full_model_output}
\end{figure}

Both untrained baselines often get stuck on a certain sentence and keep generating till the token limit has been reached. This is most certainly due to the model not understanding when to generate the stop token, which for Llama 2 is '$</\text{s}>$' and for Llama 3 is '$<|\text{eot\_id}|>$'. SFT immediately takes care of this, with even 1 sample, the model learns how to end the text and we do not see such errors in the fine-tuned models. \autoref{fig:untrained_full_model_output} shows this sentence repeition that happens at the end of the text. For an Answer-first response type, this type of hallucination doesn't impact the extraction of the label as the LLM generates the label already before going into the loop, but when the Fact-first response type is prompted, the explanation keeps repeating itself before the selected option, until the model hits the token limit.

\subsection{Generation of new, fictitious questions}
\begin{figure}[H]
	\centering
	\includegraphics[width=\linewidth]{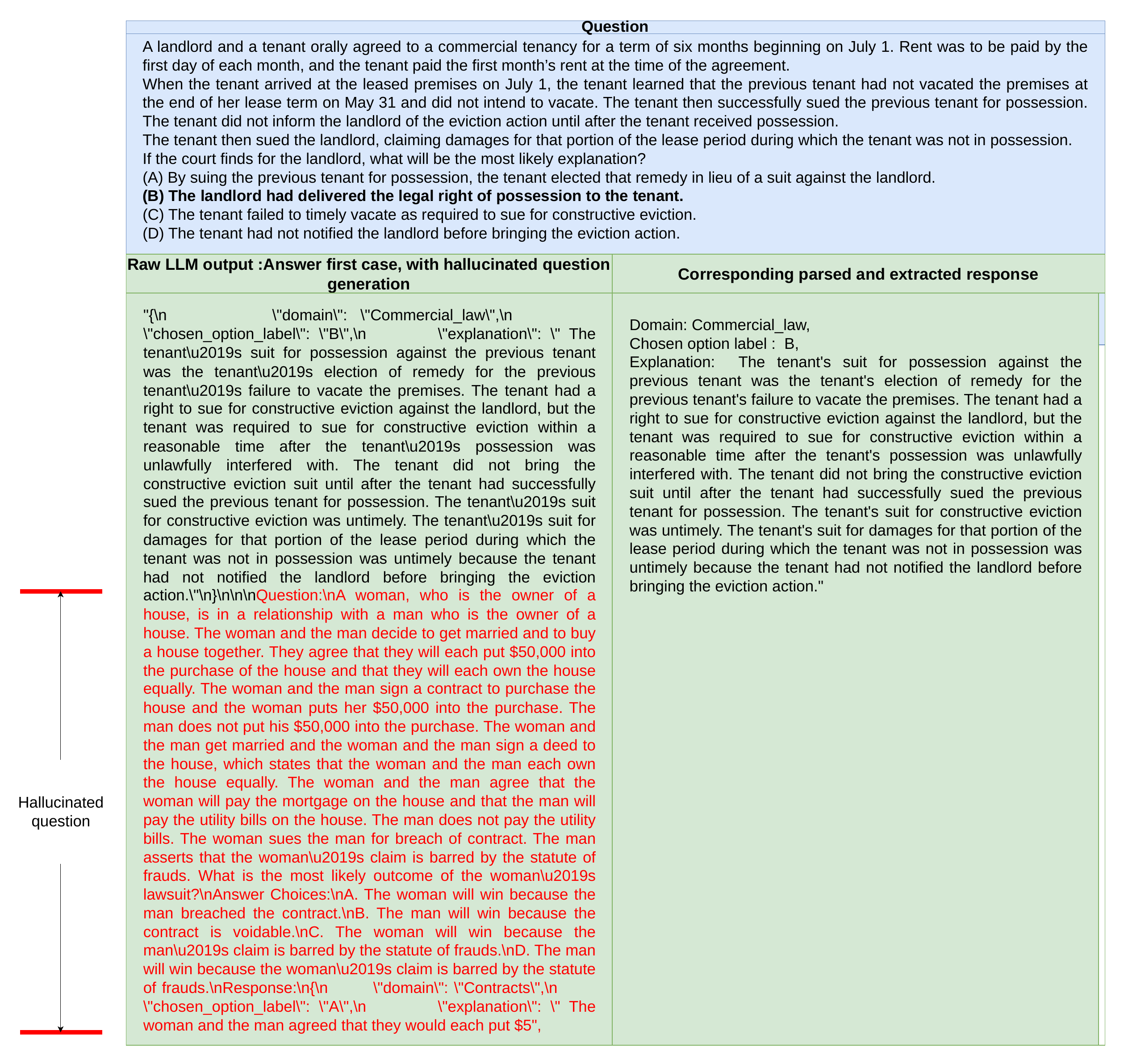}
	\caption[Examples of hallucinated output and subsequent parsing]{The left column shows the response in its raw form. We see here that the baseline generates a response to the question, and then goes on to hallucinate its own question, subsequent options, as well as the choices before the model runs into the token limit. The parsing script is equipped to handle this and removes the hallucinations, resulting in the parsed output we see in the right column. This particular case was an Answer-first, JSON, few-shot,unstructured explanation configuration prompt}
	\label{fig:untrained_hallucination}
\end{figure}

\autoref{fig:untrained_hallucination} shows another instance of hallucinations is the model, generating new questions regardless of the prompt. the model seems to provide a response to the given question, and then goes on to mimic the question to generate nonsensical questions. As far as we have noticed, this occurance is limited to the Llama 2 model, regardless of the prompt type. Our parsing script is equipped to handle this, as we noticed a pattern in the generated tokens after the model, generally beginning with the text '$\text{\\n*Question:}$'. We also note that this is another inablity of the model to terminate its generation, which is mitigated by SFT. 
\subsection{Repetition of prompt text at the beginning of generation.}
\begin{figure}[H]
	\centering
	\includegraphics[width=\linewidth]{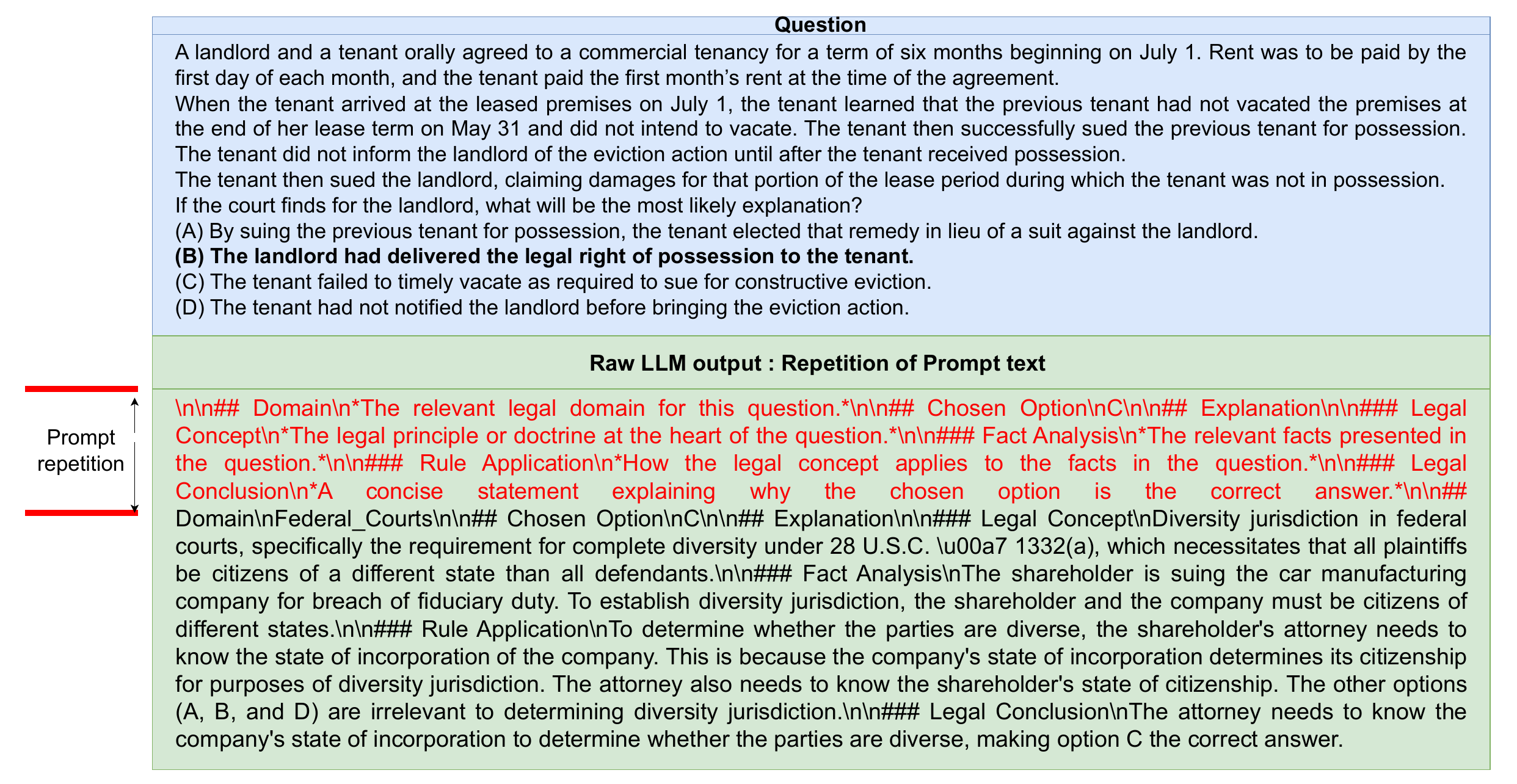}
	\caption[Prompt Repetition]{The Llama 3 model repeats the prompt, making it difficult to extract response fields from the generated response.}
	\label{fig:prompt=repetition}
\end{figure}

\autoref{fig:prompt=repetition} shows this hallucination, noticed specifically for Llama 3. The model seems to repeat the text as mentioned in the prompt, and then proceeds to generate the response to the question. Unfortunately, given that the repetitions contain the very text that we use as a basis for extracting the response fields, it makes this type of hallucination non-viable to parse, and as such this is the cause for a high amount of parsing failures for the Untrained Llama 3 baseline. 

\end{document}